\pdfoutput=1
\documentclass[letterpaper, 10 pt, conference]{ieeeconf}  % Comment this line out
\IEEEoverridecommandlockouts                              % This command is only
\usepackage{graphics} % for pdf, bitmapped graphics files
\usepackage{epsfig} % for postscript graphics files
\usepackage{mathptmx} % assumes new font selection scheme installed
\usepackage{times} % assumes new font selection scheme installed
\usepackage{amsmath} % assumes amsmath package installed
\usepackage{amssymb}  % assumes amsmath package installed

\usepackage{algorithm}
\usepackage{algorithmicx}
\usepackage{booktabs}
\usepackage{color}

\newtheorem{theorem}{Theorem}

\newtheorem{lemma}{Lemma}

\usepackage[noend]{algpseudocode}
\newcommand{\undb}[2]{\underbrace{#1}_\text{#2}}
\newcommand{\lrp}[1]{\left(#1\right)}
\newcommand{\lrs}[1]{\left[#1\right]}
\newcommand{\lrc}[1]{\left\{#1\right\}}
\newcommand{\lra}[1]{\left|#1\right|}
\newcommand{\Unif}{\textnormal{Unif}}

\newcommand{\given}{\middle\vert}
\DeclareMathOperator{\E}{\mathbb{E}}
\DeclareMathOperator*{\argmax}{arg\,max}

\newcommand{\ilim}[2]{\bigg|_{#1}^{#2}}
\newcommand{\done}{\;\blacksquare}

\usepackage{enumerate}
\usepackage{titletoc}
\title{\LARGE \bf
Voronoi Progressive Widening: Efficient Online Solvers for Continuous State, Action, and Observation POMDPs 
}

\author{Michael H. Lim, Claire J. Tomlin and Zachary N. Sunberg % <-this % stops a space
\thanks{Michael H. Lim and Claire J. Tomlin are with the Electrical Engineering and Computer Sciences department, University of California, Berkeley. Zachary N. Sunberg is with the Aerospace Engineering Sciences department, University of Colorado Boulder.}}%
\begin{document}

\maketitle
\thispagestyle{empty}
\pagestyle{empty}

%%%%%%%%%%%%%%%%%%%%%%%%%%%%%%%%%%%%%%%%%%%%%%%%%%%%%%%%%%%%%%%%%%%%%%%%%%%%%%%%
\begin{abstract}
    This paper introduces Voronoi Progressive Widening (VPW), a generalization of Voronoi optimistic optimization (VOO) and action progressive widening to partially observable Markov decision processes (POMDPs).
    Tree search algorithms can use VPW to effectively handle continuous or hybrid action spaces by efficiently balancing local and global action searching.
    This paper proposes two VPW-based algorithms and analyzes them from theoretical and simulation perspectives.
    Voronoi Optimistic Weighted Sparse Sampling (VOWSS) is a theoretical tool that justifies VPW-based online solvers, and it is the first algorithm with global convergence guarantees for continuous state, action, and observation POMDPs.
    Voronoi Optimistic Monte Carlo Planning with Observation Weighting (VOMCPOW) is a versatile and efficient algorithm that consistently outperforms state-of-the-art POMDP algorithms in several simulation experiments.
\end{abstract}

%%%%%%%%%%%%%%%%%%%%%%%%%%%%%%%%%%%%%%%%%%%%%%%%%%%%%%%%%%%%%%%%%%%%%%%%%%%%%%%%

%%%%%%%%%%%%%%%%%%%%%% Introduction %%%%%%%%%%%%%%%%%%%%%%
\section{Introduction}
The partially observable Markov decision process (POMDP) is a flexible mathematical framework for expressing stochastic sequential decision problems. 
POMDPs can represent a wide range of real world problems such as autonomous driving \cite{bai2015intention,sunberg2017value}, cancer screening \cite{ayer2012mammography}, spoken dialog systems \cite{young2013pomdp}, and aircraft collision avoidance~\cite{holand2013optimizing}.
A POMDP is an optimization problem in which we aim to find a policy that maps states to actions which will control the state to maximize the expected sum of rewards.
Finding an optimal POMDP policy is computationally demanding because of the uncertainty introduced by imperfect observations~\cite{papadimitriou1987complexity}.
One of the most popular approaches to deal with this computational challenge is to use \emph{online} algorithms that look for local approximate policies as the agent interacts with the environment rather than computing a global policy that maps every possible outcome to an action.
Many online POMDP algorithms are variants of Monte Carlo tree search (MCTS)~\cite{browne2012,Silver2010,Sunberg2017} or other tree search variants~\cite{ye2017despot,kurniawati2016online}.

\begin{table*}[t]
  \centering
  \small
  \caption{Summary of continuous space MDP and POMDP solvers studied and newly proposed in this paper.}
  \label{tab:solvers}
  \begin{tabular}{lllll}
    \toprule
    Solver & Problem & Continuous Space & Source & Brief Description \\
    \midrule
    VOOT & MDP & $S, A$ & \cite{kim2020} & VOO for deterministic MDPs; optimality guarantees; practical. \\
    POWSS & POMDP & $S, O$  & \cite{lim2020} & Sparse sampling with observation weights; optimality guarantees. \\
    \textbf{VOWSS} & POMDP & $S, A, O$ & \textbf{New} & Extends POWSS with VPW; optimality guarantees. \\
    POMCPOW & POMDP & $S, A, O$ & \cite{Sunberg2017} & Combines POMCP and DPW; practical. \\
    \textbf{VOMCPOW} & POMDP & $S, A, O$ & \textbf{New} & Extends POMCPOW with VPW; practical. \\
    BOMCP & POMDP & $S, A, O$ & \cite{mern2020} & Extends POMCPOW with Gaussian processes; practical. \\%\hline\hline
    VOSS & MDP & $S,A$ & Appendix &  Extends sparse sampling with VPW; optimality guarantees. \\
    \bottomrule
  \end{tabular}
\end{table*}

The research presented here concerns \emph{continuous space POMDPs}, defined as POMDPs with continuous state, action and observation spaces.
If tree search is naively applied to continuous space POMDPs, an immediate problem is that the branching factor of the tree will be infinite, preventing evaluation of future consequences of actions deep in the tree.
This problem has been thoroughly studied, and the most popular solutions are sparse sampling~\cite{Kearns2002,lim2020,Garg2019} and progressive widening (PW)~\cite{couetoux2011double,Sunberg2017}, which limit the branching factor by only considering a randomly-selected subset of the states, actions and observations. For continuous states and observations, this random sampling is used to approximate the expectation of the next-step value in Bellman's equation. Since the Monte Carlo integration needed to approximate these expectations has straightforward and robust convergence guarantees, sparse sampling and PW are sufficient.

Continuous action spaces are, however, much more difficult to accommodate.
Instead of simply estimating expectations, planning in a problem with a continuous action space requires solving a nonconvex optimization problem at each node in the tree.
PW has been applied to these problems~\cite{couetoux2011double,Sunberg2017}, but, since PW considers a randomly sampled set of actions, it wastes a large amount of computation on suboptimal parts of the action space and has only been demonstrated on small action spaces such as one-dimensional intervals of real numbers.
Voronoi optimistic optimization (VOO) \cite{kim2020} is an approach for sequential decision problems with deterministic and fully-observable dynamics that attempts to focus computation on more promising parts of the action space by partitioning it into Voronoi cells and sampling from cells corresponding to previously successful actions.

\begin{figure}[t]
    \centering
    \includegraphics[width=\columnwidth]{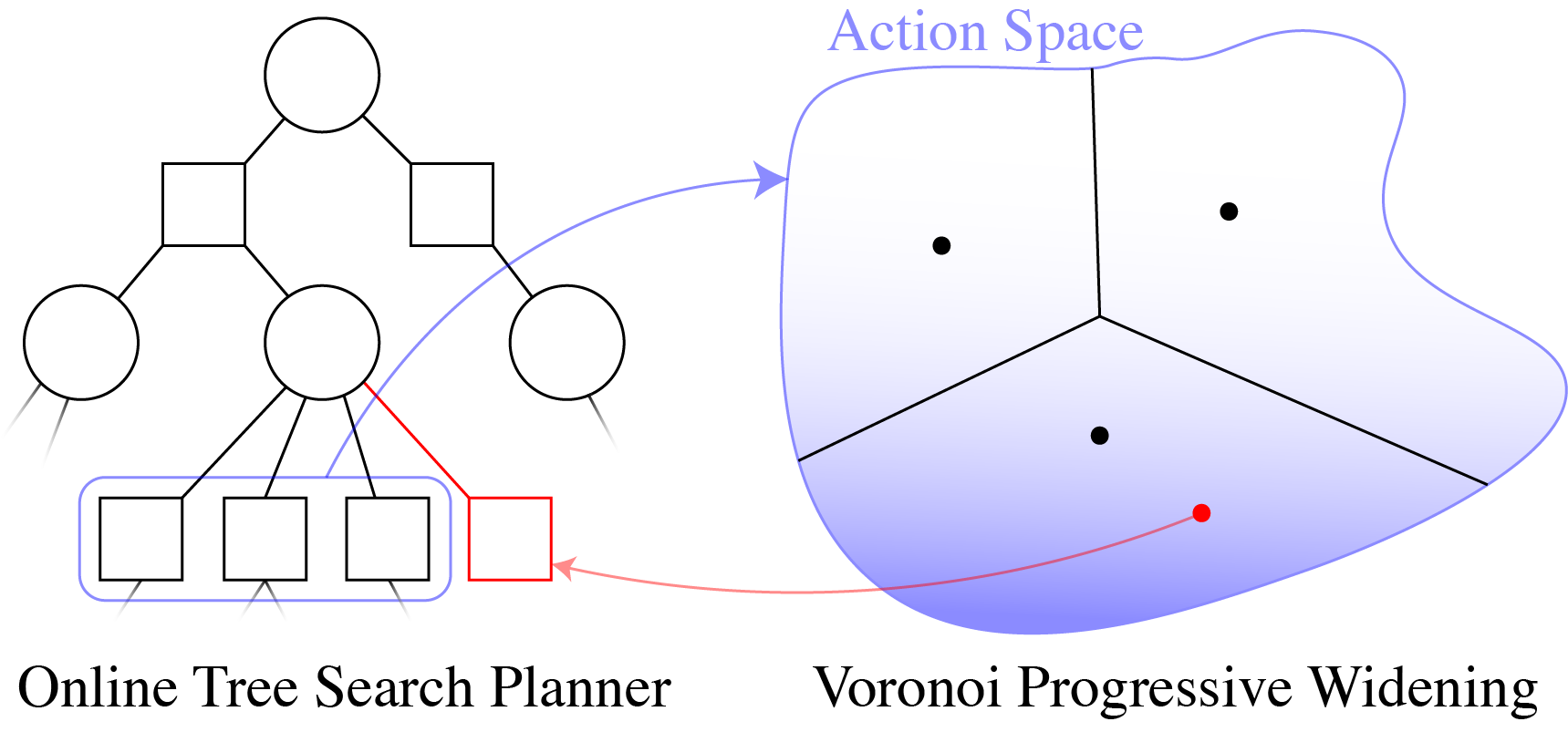}
    \caption{Voronoi progressive widening (VPW) guides the selection of actions to widen the search tree. Circles denote state/belief nodes, squares denote action nodes.}
    \label{fig:vpw-visual}
\end{figure}

In this work, we propose Voronoi Progressive Widening (VPW), a versatile technique to modify tree search algorithms to effectively handle continuous or hybrid action spaces.
VPW generalizes VOO and PW to problems with both transition and observation uncertainties. 
Like PW, it balances exploring previously proposed actions and searching for new action candidates.
However, VPW searches for new candidates in a much more efficient way via VOO, and balances local and global searching without relying on any additional prior information or expert knowledge.
Furthermore, VPW does not require significant additional computation time and can handle both continuous and hybrid action spaces with relative ease.

There are two main contributions in this work.
First, we prove theoretical guarantees about certain parts of the VPW approach via Voronoi Optimistic Weighted Sparse Sampling (VOWSS).
VOWSS is special case of VPW applied to Partially Observable Weighted Sparse Sampling (POWSS) \cite{lim2020} with fixed number of action samples, which integrates the two convergence guarantees from VOO and POWSS into a single guarantee. 
Specifically, this is the first known solver to have global convergence guarantees for continuous space POMDPs. 
This shows that VOO and action PW were extended in a way that VPW not only can handle transition and observation uncertainties, but also combines the two procedures in a theoretically sound way.

Second, we propose an efficient VPW-based algorithm: Voronoi Optimistic Monte Carlo Planning with Observation Weighting (VOMCPOW). 
VOMCPOW can also be thought of as a practical extension of VOWSS.
The experiments over different domains show the practical effectiveness of VPW for handling continuous and hybrid action space problems.
VOMCPOW consistently outperforms state-of-the-art continuous space POMDP solvers based on PW, such as Partially Observable Monte Carlo Planning with Observation Widening (POMCPOW) \cite{Sunberg2017} and Bayesian Optimized Monte Carlo Planning (BOMCP) \cite{mern2020} in several simulation experiments.
Table \ref{tab:solvers} summarizes the solvers that are studied in this paper.
    
The remainder of this paper proceeds as follows:
First, Sections \ref{sec:prelim} and \ref{sec:related} review preliminary definitions and previous work.
Then, Section \ref{sec:vpw} introduces the VPW algorithm, and describes its strengths as well as how to efficiently implement it in practice.
Section \ref{sec:theory} presents the VOWSS algorithm and theoretical analysis justifying VPW.
Finally, Section \ref{sec:experiments} empirically shows the optimality of VOWSS action selection and the efficiency and robustness of VOMCPOW over three different simulation experiments.

\setlength{\textfloatsep}{5pt}

%%%%%%%%%%%%%%%%%%%%%% Preliminaries %%%%%%%%%%%%%%%%%%%%%%
\section{Preliminaries} \label{sec:prelim}
\subsubsection{POMDP Formulation}
A POMDP is defined by a 7-tuple $(S, A, O, \mathcal{T}, \mathcal{Z}, R, \gamma)$: $S$ is the state space, $A$ is the action space, $O$ is the observation space, $\mathcal{T}$ is the transition density $\mathcal{T}(s'|s,a)$, $\mathcal{Z}$ is the observation density $\mathcal{Z}(o|a,s')$, $R$ is the reward function, and $\gamma \in [0,1)$ is the discount factor \cite{kochenderfer2015decision,bertsekas2005dynamic}. 
For POMDPs, since the agent receives only observations, the agent can infer the state by maintaining a belief $b_t$ at each step $t$ and updating it with new action and observation pair $(a_{t+1},o_{t+1})$ via Bayesian filtering \cite{kaelbling1998planning}. 
A policy, denoted with $\pi$, maps beliefs $b_t$ to actions $a_t$. 
The agent seeks an optimal policy, $\pi^*$ that maximizes the expected cumulative reward.

For simplicity, in the theoretical portion of this paper, we focus on problems with a finite horizon length $D$, 
though the basic concepts empirically work well in the infinite horizon case.
The state value function $V$ and action value function $Q$ for a given belief $b$ and policy $\pi$ at step $t$ by Bellman updates for $t \in [0,D-1]$ are defined as follows:
\begin{align}
  V_t^\pi(b) &= \E[\sum_{i=t}^{D-1} \gamma^{i-t} R(s_i, \pi(s_i)) | b],\; V_D^\pi(b) = 0,\\
  Q_t^\pi(b,a) &= \E\lrs{R(s,a) + \gamma V_{t+1}^\pi(bao)| b} \text{,}
\end{align}
where $bao$ indicates the belief $b$ updated with $(a,o)$.
The optimal value functions at each depth $t$ should satisfy
\begin{align}
  V_t^\star(b) &= \max_{a \in A}Q_t^\star(b,a),\; \pi_t^\star(b) = \argmax_{a\in A}Q_t^\star(b,a),\\
  Q_t^\star(b,a) &= \E\lrs{R(s,a) + \gamma V_{t+1}^\star(bao)| b}.
\end{align}

\subsubsection{Generative models} 
In some cases, it is not necessary to explicitly evaluate the probability density of the transition or observation distributions, and merely generating samples is sufficient. 
In this work, we use a \emph{generative model} $G$ that generates state, reward, and observation samples.

%%%%%%%%%%%%%%%%%%%%%% Related Work %%%%%%%%%%%%%%%%%%%%%%
\section{Additional Related Work} \label{sec:related}
This section expands on the introduction to cover more previous work in solving continuous action MDPs and POMDPs with online tree search.
Several techniques use double progressive widening (DPW) \cite{couetoux2011double}, originally designed to solve continuous space MDPs.
Most notably, POMCPOW and PFT-DPW \cite{Sunberg2017} extend POMCP and DPW to handle continuous space POMDPs. 
However, these algorithms use DPW as it is proposed with the inefficient action sampling that is not suitable for large problems.

One effective direction to handle continuous action spaces has been to use continuous bandits to sample new actions.
Particularly, hierarchical optimistic optimization (HOO) and the corresponding HOOT algorithm \cite{mansley2011} began work on continuous bandit algorithms applied to MCTS. 
This is followed by works such as HOLOP \cite{weinstein2012} that plans with open-loop trajectories instead of individual actions using HOO, POLY-HOOT \cite{mao2020} with polynomial guarantees, and VOO and VOOT \cite{kim2020} that we build on in this work.

Another direction is to use Bayesian optimization to efficiently sample new actions.
CBTS \cite{morere2016} uses Gaussian processes (GP) to tackle the random action sampling problem. 
However, CBTS does not use progressive widening approach and uses a separate GP at each belief node, which limits its optimization scope and branching capabilities.
Most recently, BOMCP \cite{mern2020} extends POMCPOW by posing the random action sampling step of DPW as a Bayesian optimization problem over all the belief and action nodes.
Despite the effectiveness of BOMCP, GPs are very computationally expensive to fit especially over the joint belief and action space, and BOMCP trades off computation time for sample efficiency compared to POMCPOW.

Other directions include GPS-ABT \cite{seiler2015} that uses generalized pattern search to find local optima, PA-POMCPOW that additionally incorporates score functions \cite{mern2020improved}, and VG-UCT that calculates gradient of the value function \cite{lee2020}.

%%%%%%%%%%%%%%%%%%%%%% VPW %%%%%%%%%%%%%%%%%%%%%%
\section{Voronoi Progressive Widening} \label{sec:vpw}

%% VOO agent
\begin{algorithm}[t]
\caption{VOO Algorithms \cite{kim2020}}\label{alg:voo}
\textbf{Global Variables:} $A,D(\cdot,\cdot), \omega, \Sigma$. \\
\textbf{Algorithm:} \textsc{VOO}($\mathbf{a}, \mathbf{Q}$)\\
\textbf{Input:} Array of Voronoi centers $\mathbf{a}= [a_i]$ and their function values $\mathbf{Q}= [Q(a_i)]$.\\
\textbf{Output:} A sample $a$.
\begin{algorithmic}[1]
    \State $u \leftarrow \Unif[0,1]$
    \If{$u \leq \omega$ or $|\mathbf{Q}| = 0$}
        \State $a \leftarrow \Unif(A)$
    \Else
        \State $a \leftarrow \textsc{BestVoronoiCell}(\mathbf{a}, \mathbf{Q})$
    \EndIf
   \State \Return $a$
\end{algorithmic}

\textbf{Algorithm:} $\textsc{BestVoronoiCell}(\mathbf{a}, \mathbf{Q})$\\
\textbf{Input:} Array of Voronoi centers $\mathbf{a}= [a_i]$ and their function values $\mathbf{Q}= [Q(a_i)]$.\\
\textbf{Output:} A sample $a$.
\begin{algorithmic}[1]
\State $a^* \leftarrow \argmax_{\mathbf{a}} \mathbf{Q}$
\While{$a$ is not closest} 
    \State $a \leftarrow N(a^*,\Sigma)$
    \State Check if  $D(a,a^*) \leq D(a,a_i)\; \forall a_i \in \mathbf{a},\; a_i \neq a^*$
\EndWhile
\State \Return $a$
\end{algorithmic}
\end{algorithm}

In this section, we first introduce VOO and PW to motivate the formulation of VPW and VOMCPOW.

\subsection{Voronoi Optimistic Optimization}
Voronoi optimistic optimization (VOO) \cite{kim2020} is a continuous multi-armed bandit algorithm that adaptively explores the sampling space by partitioning the space into Voronoi cells.
We define a Voronoi cell by the set of points closest to the corresponding Voronoi center compared to the other centers, and the best Voronoi cell is the Voronoi cell with the center that achieves the highest function value estimate.
In our setting, the sampling space is the action space, the Voronoi centers are the sampled actions, and the function values are the $Q$-value estimates at the actions.
Since Voronoi cells are solely defined by distances to Voronoi centers, \textsc{VOO} additionally takes in a distance metric $D(\cdot,\cdot)$ as an input, and scales well to higher dimensions unlike HOO.

In Algorithm \ref{alg:voo}, we define the adapted functions \textsc{VOO} and \textsc{BestVoronoiCell}.
\textsc{VOO} searches the action space either uniformly with probability $\omega$ or from the best Voronoi cell with probability $1-\omega$. 
\textsc{BestVoronoiCell} samples an action from the best Voronoi cell via rejection sampling.
In practice, VOO sampling is best implemented by using Gaussian rejection sampling centered around the current best action \cite{kim2020}.
While we reflected this in our algorithm definition, we note that the theoretical guarantees only hold for uniform rejection sampling.
Furthermore, to improve computation time for the experiments in Section~\ref{sec:experiments}, we limit the maximum number of rejected samples via methods described in Appendix B.
VOO has previously been extended to Voronoi Optimistic Optimization Tree (VOOT) \cite{kim2020}, but only for deterministic MDPs.

\subsection{Progressive Widening}

\begin{algorithm}[t]
\caption{Action Progressive Widening \cite{couetoux2011double}} \label{alg:pw}
\textbf{Global Variables:} $k_a, \alpha_a, A, c$.\\
\textbf{Algorithm:} $\textsc{PW}(h)$\\
\textbf{Input:} Belief/history node $h$ in the MCTS tree.\\
\textbf{Output:} An action $a$.
\begin{algorithmic}[1]
    \If{$|C(h)| = 0$ or $|C(h)|\leq k_aN(h)^{\alpha_a}$}
        \State $a \leftarrow \Unif(A)$ \label{line:new}
        \State $C(h) \leftarrow C(h) \cup \{a\}$
    \Else
        \State $a \leftarrow \argmax_{a\in C(h)} Q(h,a) + c\sqrt{\frac{\log N(h)}{N(h,a)}}$ \label{line:ucb}
    \EndIf
    \State \Return $a$
\end{algorithmic}
\end{algorithm}

Progressive widening (PW) \cite{couetoux2011double} tackles continuous state and action spaces by gradually expanding the state and action sample sets.
PW limits the number of sampled children to $k\cdot N ^\alpha$, where $N$ corresponds to the number of visits to the parent node in the search tree, and $k,\alpha$ are the widening parameters.
Algorithm \ref{alg:pw} describes action progressive widening.
Here, $h$ is the belief/history node in the MCTS tree, $C(h)$ the list of children action nodes, $N$ the number of visits, and $c$ the Upper Confidence Bound exploration parameter.
In the action space, PW balances two modes of exploration: (1) exploring branches with previously proposed actions (line~\ref{line:new}), and (2) searching for new suitable action candidates (line~\ref{line:ucb}).
However, since PW samples new actions uniformly from the action space, it is rather sample inefficient and does not take into account any information gained from previous samples.

\subsection{Voronoi Progressive Widening}

\begin{algorithm}[t]
\caption{Voronoi Progressive Widening} \label{alg:vpw}
\textbf{Global Variables:} $k_a, \alpha_a, \omega, A, c$.\\
\textbf{Algorithm:} $\textsc{VPW}(h)$\\
\textbf{Input:} Belief/history node $h$ in the MCTS tree.\\
\textbf{Output:} An action $a$.
\begin{algorithmic}[1]
    \If{$|C(h)| = 0$ or $|C(h)|\leq k_aN(h)^{\alpha_a}$}
        \State $a \leftarrow {\color{red}\textsc{VOO}(C(h), [Q(h,\cdot)])}$
        \State $C(h) \leftarrow C(h) \cup \{a\}$
    \Else
        \State $a \leftarrow \argmax_{a\in C(h)} Q(h,a) + c\sqrt{\frac{\log N(h)}{N(h,a)}}$
    \EndIf
    \State \Return $a$
\end{algorithmic}
\end{algorithm}

We now introduce Voronoi Progressive Widening (VPW) in Algorithm \ref{alg:vpw}, which generalizes VOO and PW. 
Essentially, VPW progressively widens with a VOO sample instead of a uniform sample. 
With this formulation, PW is simply VPW with $\omega =1$, and VOO is VPW with $k_a\cdot N ^{\alpha_a} = +\infty$.

Compared to PW, VPW only additionally relies on having a collection of $Q$-value estimates, which are typically calculated and stored for MCTS solvers.
While this means that we require the $Q$-value estimates to be relatively faithful to the actual $Q$-values and also requires an informative rollout policy that can guide the solver to more optimal actions, this is not a requirement specific to VPW as a typical MCTS solver should aim to have both of these components.
Furthermore, since the VOO step does not require a significant additional time, VPW operates in a similar time scale as PW/DPW while being able to efficiently sample action candidates that are closer to the optimal action.
The only other requirement of VPW is the distance metric on the action space, but this is often simple to define, and in fact lets VPW to tackle hybrid action spaces such as the one in Section \ref{sec:vdp}.

\subsection{VOMCPOW}
The VPW technique is versatile as it can be applied to any MCTS solver for MDPs and POMDPs without requiring any additional expert or domain knowledge.
For instance, one could consider Sparse-UCT-VPW algorithm for continuous space MDPs that uses VPW criterion with the Sparse-UCT algorithm \cite{bjarnason2009lower} to handle continuous action spaces. 
Specifically, we demonstrate the versatility of VPW through modifying POMCPOW into VOMCPOW (Voronoi Optimistic Monte Carlo Planning with Observation Weighting) by simply swapping out action PW with VPW. 
VOMCPOW consistently outperforms POMCPOW and BOMCP by a statistically significant margin across all of our experiments.

%%%%%%%%%%%%%%%%%%%%%% VOWSS %%%%%%%%%%%%%%%%%%%%%%
\section{Convergence Analysis} \label{sec:theory}
For convergence analysis, we introduce and Voronoi Optimistic Weighted Sparse Sampling (VOWSS), a VPW-based continuous space online POMDP solver that guarantees convergence to an optimal policy. 
VOWSS justifies the usage of VPW-based solvers that build upon sparse sampling \cite{Kearns2002}, particle weighting and VOO, such as VOMCPOW.

\subsection{VOWSS} \label{sec:algorithms}

%% VOWSS
\begin{algorithm}[t]
\caption{Value estimation algorithms for VOWSS} \label{alg:vowss}
\textbf{Global Variables:} $\gamma,G,C_s,C_a,D$.\\
\textbf{Algorithm:} \textsc{EstimateV}($\bar{b}, d$)\\
\textbf{Input:} Belief particle set $\bar{b}=\{(s_i,w_i)\}$, depth $d$.\\
\textbf{Output:} A scalar $\hat{V}_d^\star(\bar{b})$ that is an estimate of $V^\star_d(\bar{b})$.
\begin{algorithmic}[1]
\State If $d \geq D$ the max depth, then return 0.
\State For $i = 1,\cdots, C_a$, sequentially run \textsc{VPW}($\bar{b}$) to sample actions $a_i$ and estimate the corresponding $Q$-values with $\hat{Q}_d^\star(\bar{b},a_i) = \textsc{EstimateQ}(\bar{b}, a, d$).
\State Return the value function:
$$\hat{V}_d^\star(\bar{b}) = \max_{i=1,\cdots,C_a} \hat{Q}_d^\star(\bar{b},a_i).$$
\end{algorithmic}

\textbf{Algorithm:} \textsc{EstimateQ}($\bar{b}, a, d$)\\
\textbf{Input:} Belief particle set $\bar{b}=\{(s_i,w_i)\}$, action $a$, depth $d$.\\
\textbf{Output:} A scalar $\hat{Q}_d^\star(\bar{b},a)$ that is an estimate of $Q_d^\star(b,a)$.
\begin{algorithmic}[1]
\State For each particle-weight pair $(s_i,w_i)$ in $\bar{b}$, generate $s_i',o_i,r_i$ from $G(s_i,a)$.
\State For each observation $o_j$ from previous step, iterate over $i=1,\cdots,C_s$ to insert $(s_i', w_i \cdot \mathcal{Z}(o_j|a,s_i'))$ to a new belief particle set $\overline{bao_j}$.
\State Return the $Q$-value estimate:
\end{algorithmic}
\begin{align*}
  &\hat{Q}_d^\star(\bar{b},a) = \frac{\sum_{i = 1}^{C_s} w_i(r_i+\gamma\cdot \textsc{EstimateV}(\overline{bao_i}, d+1))}{\sum_{i = 1}^{C_s} w_i} .
\end{align*}
\end{algorithm}

We define \textsc{EstimateV} and \textsc{EstimateQ} functions in Algorithm \ref{alg:vowss}.
Here, $C_s,C_a$ are the state and action widths, respectively.
\textsc{EstimateV} is a subroutine that returns the value function, $V$, for an estimated state or belief, by calling \textsc{EstimateQ} for each action and returning the maximum.
Similarly, \textsc{EstimateQ} performs sampling and recursive calls to \textsc{EstimateV} to estimate the $Q$-function at a given step with a weighted average in the style of POWSS~\cite{lim2020}. 
The sampled states $s'_i$ are inserted into each next-step belief particle set $\overline{bao_j}$ with the new weights $w_i' = w_i \cdot \mathcal{Z}(o_j|a,s_i')$, which are the adjusted probability of hypothetically sampling observation $o_j$ from state $s_i'$.

Thus, the VOWSS policy action is obtained by calling \textsc{EstimateV}$(\bar{b}_0,0)$ at the root node and taking the action that corresponds to the maximizing $Q$-value.
Note that the particle belief set is initialized by drawing samples from $b_0$ and normalizing weights to $1/C_s$.
Like POWSS, VOWSS is not very computationally efficient as it fully expands the sparsely sampled tree.
Thus, VOWSS mainly demonstrates theoretical convergence and are only practically applicable to very small problems.
The algorithms in this section are written in a style closer to a plain English format to enhance the readability of mathematical machinery used for these algorithms, since VOWSS is more theoretical in nature.

It is worth noting that with our recursive formulation, VOOT \cite{kim2020} can be recast as a $Q$-function estimation algorithm that simply returns the one sample $Q$-function estimate. 
Particularly, this means that \textsc{EstimateQ} for VOOT would look identical to VOWSS's \textsc{EstimateQ} function except with $C_s$ set to 1 as VOOT assumes no transition and observation uncertainties, with no action width decay.
Consequently, VOOT is notably faster than VOWSS and thus practical for deterministic MDP problems, since it only needs to sample the deterministic next step state once.
Nevertheless, due to the hierarchical structure of VOOT, the actual VOO algorithm as well as the theoretical guarantees of VOOT should remain identical when recast into this recursive form.
We also discuss the stochastic MDP case in Appendix C.

\subsection{Convergence Guarantees} \label{sec:guarantees}
We obtain global convergence to the optimal value functions for continuous space POMDPs by combining VOO that globally optimizes over the entire action space and POWSS that estimates value functions with arbitrary precision.
To our knowledge, VOWSS is the first continuous space POMDP algorithm to have global convergence guarantees without relying on any discretization schemes for any of the spaces.
In the subsequent sections, we prove that VOWSS policy can be made to perform arbitrarily close to the optimal policy by increasing both the state and action widths.
Our results are derived with $k_a\cdot N ^{\alpha_a} = +\infty$ (VPW = VOO), and uniform rejection sampling for \textsc{BestVoronoiCell}.

The convergence proof of VOWSS relies on the proofs for the ancestor algorithms: POWSS and VOO.
Since the proof of VOWSS requires the union of all the regularity conditions for POWSS and VOO, which deal with local smoothness of rewards and observation density requirements, we will not repeat them here and detail them in Appendix A.2.
One of the conditions requires that the reward function $R$ be bounded and measurable, $||R||_{\infty} \leq R_{\max}$, and consequently the value is bounded by $V_{\max} \equiv R_{\max}/(1-\gamma)$ for $\gamma < 1$. 

\begin{theorem}[VOWSS Inequality] \label{thm:vowss}
    Suppose we choose the action sampling width $C_a$ and state sampling width $C_s$ such that under the union of regularity conditions specified by \cite{lim2020} and \cite{kim2020}, the intermediate POWSS bounds and VOO bounds in Lemma \ref{lemm:vowss} are satisfied at every depth of the tree. Then, the following bounds for the VOWSS estimator $\hat{V}_{\textsc{VOWSS},d}(\bar{b})$ hold for all $d\in[0,D-1]$ in expectation:
    \begin{align}\label{eq:vowss}
        |V_d^\star(b) - \hat{V}_{\textsc{VOWSS},d}(\bar{b})| &\leq \eta_{C_a} + \alpha_{C_s}.
    \end{align}
\end{theorem}

Here, $\eta_{C_a}$ and $\alpha_{C_s}$ are non-negative bounds that tend to 0 as we increase $C_a$ and $C_s$, respectively. We show a brief outline of the proof of Theorem \ref{thm:vowss}.
In order to prove Theorem \ref{thm:vowss}, we first need to prove the intermediate bounds. 

\begin{lemma}[VOWSS Intermediate Inequality] \label{lemm:vowss}
  Suppose with our notation, the POWSS estimators at all depths $d$ are within $\epsilon$ of their mean values with probability $1-p$, and the VOO agents have regret bounds of $\mathcal{R}_{C_a}$.
  The following inequalities hold for all $d\in[0,D-1]$ in expectation:
  \begin{align}
    |V_d^\star(b) - \hat{V}_d^{C_a}(b)| &\leq \eta_{C_a}(d),\\
    |\hat{V}_d^{C_a}(b) - \hat{V}_{\textsc{VOWSS},d}(\bar{b})| &\leq \alpha_{C_s}(d).
  \end{align}
\end{lemma}

In Lemma \ref{lemm:vowss}, we aim to bound the inequality in (\ref{eq:vowss}) by applying triangle inequality to the two split terms: the VOO-like regret bound with $\eta_{C_a}(d)$ and the POWSS-like concentration bound with $\alpha_{C_s}(d)$.
The exact definitions of these intermediate bound terms are defined and explained further in Appendix A.2.
We define each value function used in Lemma \ref{lemm:vowss} as the following:
\begin{align}
  &V_d^\star(b)\equiv \max_{a\in A} \; R(b,a) + \gamma \E[V^\star_{d+1}(bao)|b] \notag,\\
  &\hat{V}_d^{C_a}(b) \equiv \max_{a\in VOO(A)} R(b,a) + \gamma \E[\hat{V}_{d+1}^{C_a}(bao)|b],\\
  &\hat{V}_{\textsc{VOWSS},d}(\bar{b}) \equiv \max_{a\in VOO(A)} \frac{\sum_{i = 1}^{C_s} w_i(r_i+\gamma\cdot \hat{V}_{\textsc{VOWSS},d+1}(\overline{bao_i}))}{\sum_{i = 1}^{C_s} w_i}.  \notag
\end{align}
Here, $V_d^\star(b)$ is the optimal value function as per the conventional definition, and $\hat{V}_{\textsc{VOWSS},d}(\bar{b})$ is the VOWSS estimator for the optimal value function.
In addition, we introduce the theoretical intermediate term $\hat{V}_d^{C_a}(b)$ that bridges the gap between the regret bound and the concentration bound by only performing the VOO step and not the sparse sampling step.
Using this intermediate term, we can further subdivide the intermediate inequalities to obtain the appropriate regret and concentration bounds using results from \cite{kim2020} and \cite{lim2020}, respectively.
Proving Lemma \ref{lemm:vowss} proves Theorem \ref{thm:vowss}.

%%%%%%%%%%%%%%%%%%%%%% Experiments %%%%%%%%%%%%%%%%%%%%%%
\section{Experiments} \label{sec:experiments}
The numerical experiments in this section confirm the theoretical results of Section \ref{sec:guarantees}, as well as showcase the effectiveness of VPW. 
We demonstrate the performances of our algorithms in three different control experiments: Linear-Quadratic-Gaussian (LQG) control, Van Der Pol Tag, and modified lunar lander problems.
When running experiments for POMCPOW, VOMCPOW, and BOMCP, we use the rollout policy heuristic to estimate the value function as well as take the first action to be the rollout policy action at each newly generated belief node.
To determine the hyperparameters of the solvers, we used cross-entropy method (CEM) \cite{mannor2003} to maximize the mean expected reward of a hyperparameter set when the optimal hyperparameters are not already available from previous experiments.
The VOMCPOW and BOMCP hyperparameters were trained by first initializing them with POMCPOW hyperparameters and then using CEM again including all the additional hyperparameters specific to the solvers.
The only hyperparameter that was manually chosen is the Gaussian covariance matrix for VOO rejection sampling.
The code for the experiments is built on the POMDPs.jl framework \cite{egorov2017pomdps}.
The exact hyperparameters used for each experiment are given in Appendix B.

\subsection{Linear-Quadratic-Gaussian Control} \label{sec:lqg}
We first test our algorithms on a simple 2D-action space LQG control system.
Here, we choose a relatively simple problem setup such that we can easily understand and visualize the results while allowing VOWSS to still plan within a reasonable time.
Here, VOWSS uses VPW like VOO by setting $k_a\cdot N ^{\alpha_a} = +\infty$, but with Gaussian rejection sampling.
The dynamics and observation models are
\begin{align}
  x_{t+1} &= x_t + u_t + v_t;\;\; y_{t} = x_t + w_t;\;\; v_t ,w_t\stackrel{i.i.d.}{\sim} N(0, \sigma^2 I).
\end{align}
The initial state $x_0$ is distributed as $N([-10,10], \sigma^2 I)$, and  $\sigma=0.1$ for all $x_0,v_t,w_t$.
We aim to minimize the cost function $J(x_0)$, while planning for two steps $(N=2)$:
\begin{align}
  J(x_0) &= \E[x_N^Tx_N + \sum_{t=0}^{N-1} (x_i^Tx_i + u_i^Tu_i)].
\end{align}
The analytical answer can be obtained by solving the LQG backup equation, which is identical to the LQR solution,
\begin{align}
  u_0^\star = -K_0 \hat{x}_0 = -0.6 \cdot \hat{x}_0 = [6.0,-6.0].
\end{align}

Since the problem has an analytical solution, we can directly compare the actions each solver chooses to the analytical solution, for VOWSS, POMCPOW, VOMCPOW and BOMCP. 
For the solvers that require a rollout policy, we test both the finite horizon LQR solution (referred to as ``exact policy'') and the steady-state solution to the discrete time algebraic Riccati equation (referred to as ``Riccati policy'') \cite{chow1975} as the rollout policies to observe effects on each solver. 
For this particular scenario, the Riccati solution is $u_t^* \approx -0.618\cdot\hat{x}_t$, and $u_0^* \approx [6.18,-6.18]$, slightly different from the exact policy.

\setlength{\textfloatsep}{12pt}
\begin{figure}[t]
    \centering
    \includegraphics[width=\columnwidth]{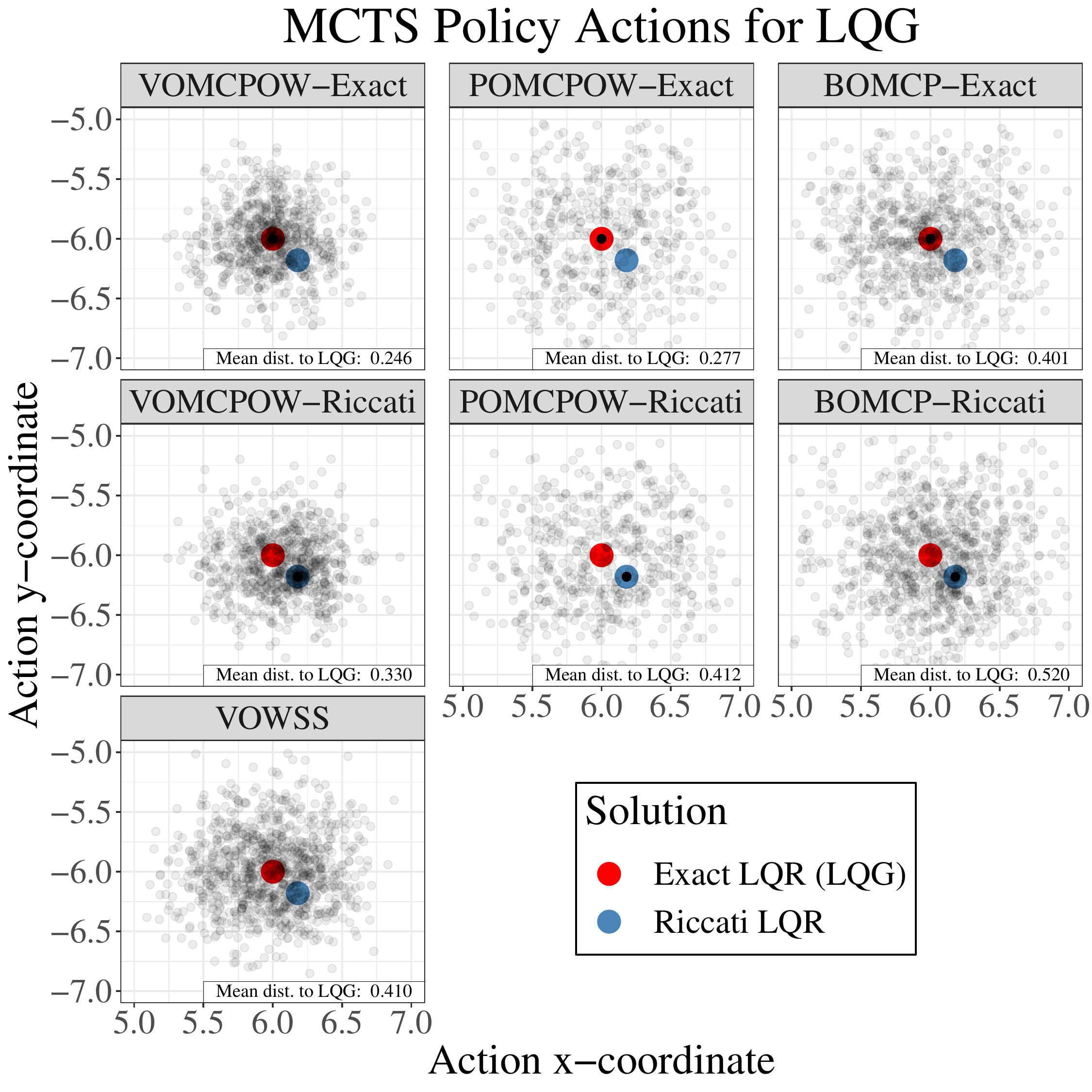}
    \caption{Scatter plots of the first actions chosen by policies from different solvers and rollouts for the two step LQG problem. The exact LQR solution, which is the same as the LQG solution, is shown as a red dot at $[6.0,-6.0]$, and the Riccati solution as a blue dot at approximately $[6.18,-6.18]$.}
    \label{fig:lqg-scatter}
\end{figure}

When choosing hyperparameters through CEM, we use the Riccati policy as the rollout policy.
The number of queries for POMCPOW and VOMCPOW are set to 1000 and BOMCP to 100, which correspond to approximately 0.1 seconds of planning time. 
For VOWSS, the state width is set to 10 and action width to 200. 
Furthermore, we introduce the action width decay term $\gamma_a$ such that the number of VPW iterations for a given height is $\gamma_a$ times the VPW iterations for the previous height, similar to the way VOOT is modified for long horizons \cite{kim2020}.
The action width decay is manually set to $\gamma_a =0.4$ after a careful inspection to balance performance and run time, which means that the action width for first step is 200, and the action width for second step is 80.
While this results in number of evaluations on the order of $10^6$, we include the performance of VOWSS as a reference.

We first show the scatter plots of the actions chosen by each solver in Figure \ref{fig:lqg-scatter}.
Each scatter plot shows the result of 1000 simulations.
While POMCPOW and BOMCP find solutions biased towards the rollout actions, we see that the effect is much less pronounced in VOMCPOW.
Specifically, when the rollout policy is set to Riccati policy, both POMCPOW and BOMCP heavily resort to picking the Riccati solution as shown by the large point mass corresponding to the Riccati solution in the scatter plots.
Even though VOMCPOW still produces a sizable mass of points on and surrounding the Riccati solution, it otherwise picks solutions that are close to the LQG solution.
VOMCPOW scatter plots show that the bias and the variance of the optimal action estimates are noticeably smaller than those of POMCPOW and BOMCP.
Although VOWSS cannot be directly compared to the other solvers since the number of iterations is not on the same order and it doesn't rely on a rollout policy, the overall shape of the scatter plot looks similar to those of VOMCPOW.

\begin{figure}[t]
    \centering
    \includegraphics[width=\columnwidth]{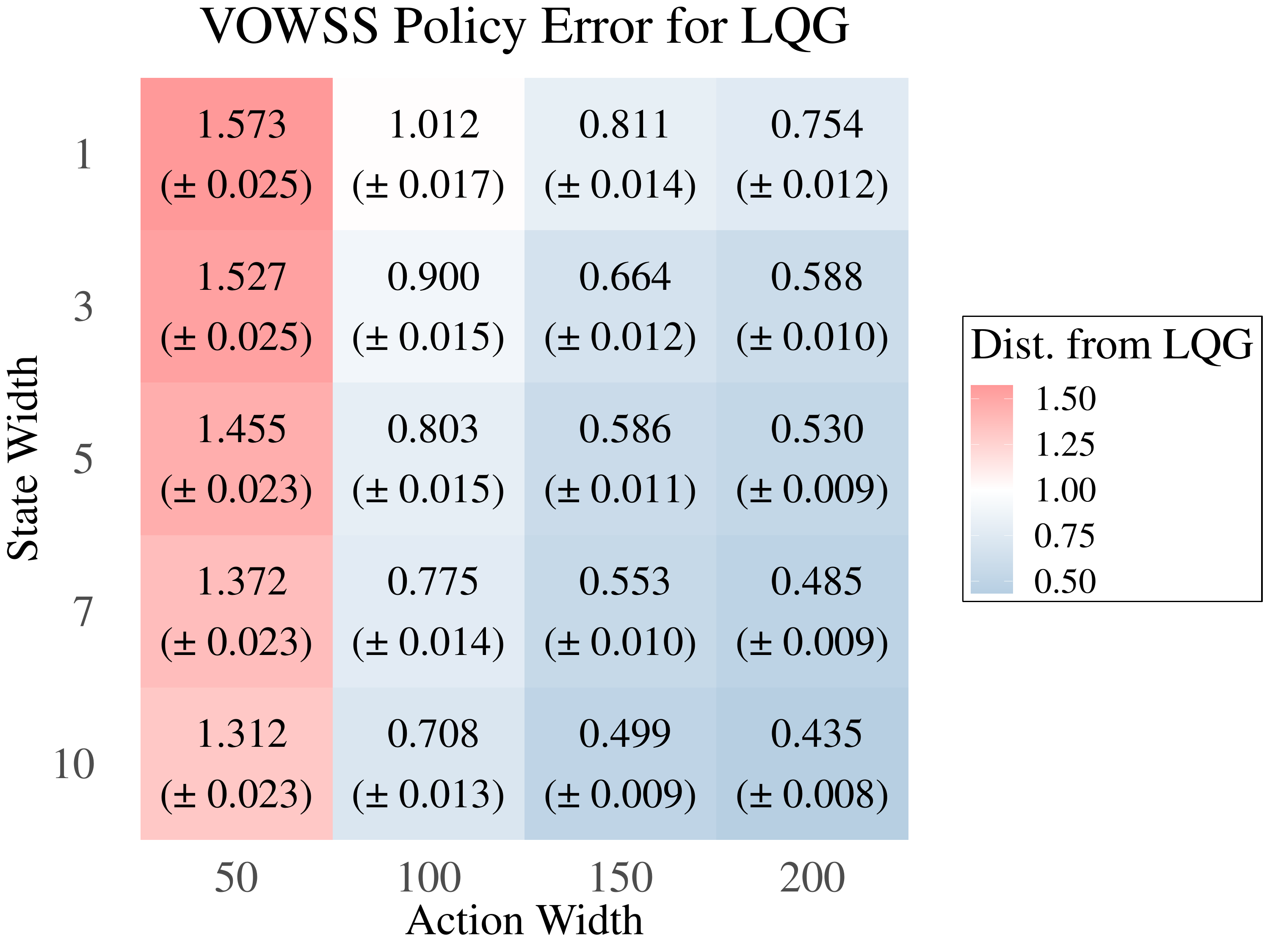}
    \caption{Tabular summary of the first actions chosen by VOWSS policy, where we show the mean Euclidean distance from the LQG solution, and the corresponding standard error in parentheses.}
    \label{fig:lqg-vowss}
\end{figure}

In addition, we study the effects of varying the state and action widths for VOWSS.
Figure \ref{fig:lqg-vowss} shows the mean Euclidean distance to the LQG solution for different combinations of state and action widths.
Each cell in the table shows the result of 1000 simulations, and the other hyperparameters are left intact.
As we increase either the state or action width, the mean distance and the standard error decrease.
This indicates that VOWSS chooses better actions by increasing the state and action widths, which confirms our theoretical results: larger state widths should increase value estimation accuracy and larger action widths should improve action optimization.

\subsection{Van Der Pol Tag} \label{sec:vdp}
\begin{figure}[t]
    \centering
    \includegraphics[width=\columnwidth]{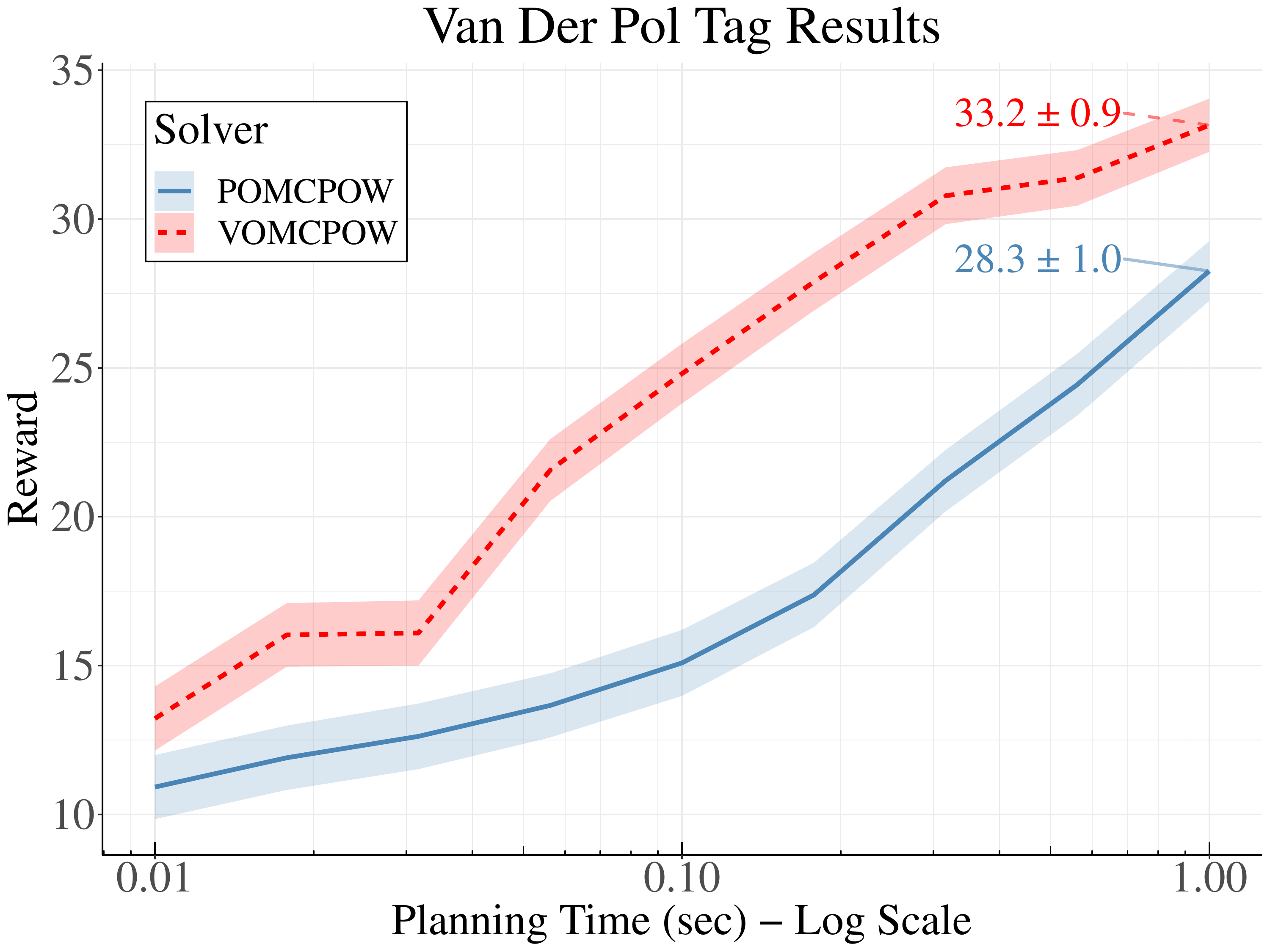}
    \caption{Mean rewards for POMCPOW and VOMCPOW for VDP Tag. Ribbons indicate one standard error.}
    \label{fig:vdp}
\end{figure}
Next, we test POMCPOW and VOMCPOW on the Van Der Pol Tag (VDP Tag) problem introduced by \cite{Sunberg2017}, which is the only problem in their work with continuous state, observation and (hybrid) action spaces.
In VDP Tag, an agent navigates through 2D box to tag a target with randomized initial state that follows a dynamics model defined by the Van Der Pol differential equation. 
The agent moves at a fixed speed, but can choose the direction of travel and can decide to make an accurate observation of where the target is at a higher cost than making the default noisy observation.
While the target can freely move within the 2D box, the agent is blocked when it comes into contact with a barrier.

We show the mean reward for 1000 simulations for each solver and each planning time plotted in log scale of seconds in Figure \ref{fig:vdp}.
We observe that VOMCPOW outperforms POMCPOW at every planning time by a statistically significant margin.
It is also worth noting that VOMCPOW takes almost an order of magnitude less planning time to reach the mean reward of 25 compared to POMCPOW.
While VDP Tag is a continuous space POMDP, the rewards are still discrete in both state and action spaces, which suggests that even with discrete jumps in the reward function, VPW can still optimize to find better actions.
In addition, while it is possible to adapt BOMCP to solve VDP Tag, it requires nontrivial modifications to the action space due to the action space being hybrid ($A = [0,2\pi) \times \{0, 1\}$) and the angle space being a modular space.
This further illustrates the ease of implementing VPW since we only need to supply VPW with a distance metric on the action space, which makes VPW suitable for hybrid action spaces as well.

\subsection{Lunar Lander} \label{sec:lander}
\begin{figure}[t]
    \centering
    \includegraphics[width=\columnwidth]{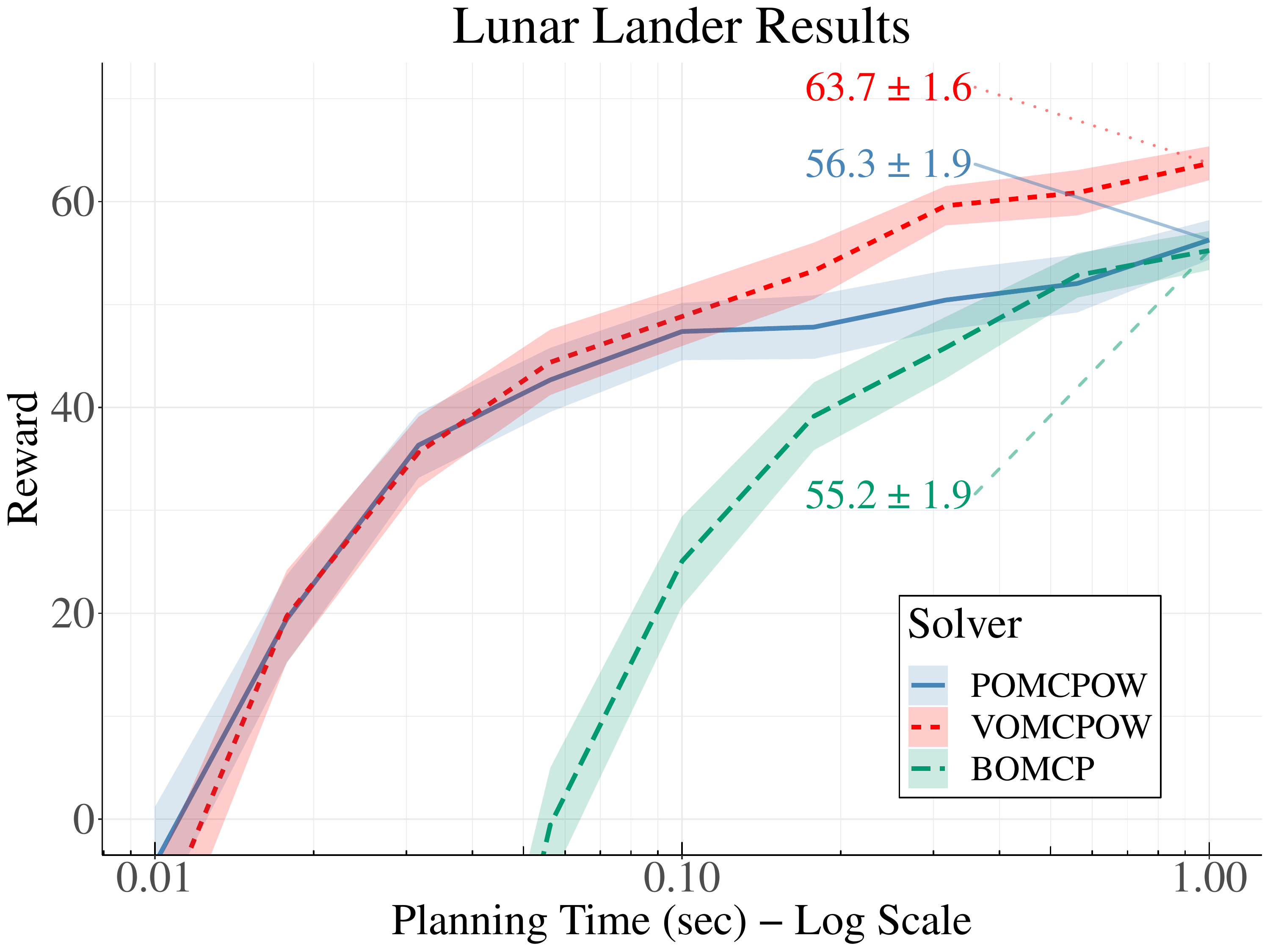}
    \caption{Mean rewards for POMCPOW, VOMCPOW and BOMCP for the modified lunar lander problem. Ribbons indicate one standard error.}
    \label{fig:lander}
\end{figure}
Lastly, we test POMCPOW, VOMCPOW and BOMCP on the modified lunar lander problem proposed in \cite{mern2020}, where the main objective is to guide a vehicle to land in a target zone safely.
In this version of lunar lander, the vehicle state is defined as $(x,y,\theta,\dot{x}, \dot{y},\omega)$, and we only obtain noisy observations of the angular rate, horizontal speed, and above-ground level.
The action space is defined by the tuple $(T,F_x,\delta)$ where $T$ is the main vertical thrust, $F_x$ the corrective horizontal thrust, and $\delta$ the offset.
The default rollout policy is proportional control based on the observation.

Once again, we show the mean reward for 1000 simulations for each solver and each planning time plotted in log scale of seconds in Figure \ref{fig:lander}.
VOMCPOW outperforms both POMCPOW and BOMCP by a statistically significant margin once the planning time exceeds 0.1 seconds. 
Below 0.1 seconds, the performances of POMCPOW and VOMCPOW are similar due to the problem having a long horizon, in which the effects of VPW will not be as apparent since best Voronoi cell actions will not be sampled very often.

We note that the performance of POMCPOW here seemingly differs from the results in \cite{mern2020}, which might be due to a several factors.
The lander problem has a high variance in the returns, since failure to land results in a steep penalty of -1000. 
This failure mode does not happen very often even within 1000 iterations, so the number of failures significantly impacts the mean rewards.
Furthermore, we are comparing the algorithms based on planning time rather than number of queries.
POMCPOW queries much faster than BOMCP, and average planning time may not sufficiently capture the relationship between total planning time and number of queries for a progressively built tree.
It is also possible that authors of \cite{mern2020} did not set POMCPOW first action to be the rollout action.

%%%%%%%%%%%%%%%%%%%%%% Conclusion %%%%%%%%%%%%%%%%%%%%%%
\section{Conclusion}
In this paper, we have introduced Voronoi Progressive Widening (VPW), a versatile technique to effectively handle continuous or hybrid action spaces in MDPs and POMDPs.
Consequently, we proposed two VPW-based tree search algorithms to demonstrate convergence guarantees and efficiency, justifying the theoretical soundness, versatility and practicality of VPW for many continuous or hybrid action space POMDPs.

This study has yielded a several key insights, which suggest some promising future directions to explore.
While each of the VPW-based algorithms either enjoy theoretical guarantees or computational efficiency, the gap between theoretical soundness and practicality still remains.
In particular, the state and action selection heuristics as it is utilized in POMCPOW still need to be integrated into the theoretical algorithms like POWSS to bridge the gap between theory and practice.
On the other hand, we believe that the VPW technique itself should also have some theoretical guarantees similar to the DPW technique \cite{couetoux2011double}.

Additionally, since the convergence guarantees of VOO hold as long as the reward function is locally smooth around at least one global optimum \cite{kim2020}, more extensive study could be done for effectiveness of VPW on MDPs and POMDPs with discrete reward structures.
Similar to how the R\'enyi divergence requirement for POWSS \cite{lim2020} could be a difficulty indicator for likelihood-weighted sparse tree solvers, the proof techniques utilized for VOO and DPW could offer insights on what continuous or hybrid action space problems are harder to solve than others.

%%%%%%%%%%%%%%%%%%%%%% Acknowledgements %%%%%%%%%%%%%%%%%%%%%%
\section*{Acknowledgements}
This material is based upon work supported by a DARPA Assured Autonomy Grant, the SRC CONIX program, NSF CPS Frontiers, the ONR Embedded Humans MURI, and the National Science Foundation Graduate Research Fellowship Program under Grant No. DGE 1752814.
Any opinions, findings, and conclusions or recommendations expressed in this material are those of the authors and do not necessarily reflect the views of any aforementioned organizations.
The authors also thank John Mern for sharing the source code for BOMCP and the lunar lander environment, and providing valuable insights as well as technical support for running and verifying these scripts.

% \small
\bibliographystyle{IEEEtran}
\bibliography{VPW-arXiv}

\newpage
%%%%%%%%%%%%%%%%%%%%%% For Appendix %%%%%%%%%%%%%%%%%%%%%%
\onecolumn
\appendix
\setcounter{theorem}{0}
\setcounter{lemma}{0}
\setcounter{equation}{0}
\setcounter{figure}{0}
\setcounter{table}{0}
\setcounter{page}{1}

\startcontents
\vskip3pt\hrule\vskip5pt \textbf{Contents}
\printcontents{}{2}{} 
\vskip3pt\hrule\vskip5pt

\subsection{Mathematical Proofs}
\subsubsection{Analogous Notations and Concepts to Previous Works}
In our work, we build upon a lot of previous works on sparse sampling, POWSS, and VOO, while adapting notations and concepts to be consistent within our own work.
Table \ref{tab:variables} explicitly connects the notations we use in our analyses that are slightly different from the notations used by \cite{Kearns2002, kim2020, lim2020}.

\begin{table}[H]
    \centering
    \small
    \caption{Summary of corresponding notations and quantities.}
    \label{tab:variables}
    \begin{tabular}{lllll}
        \toprule
        Proof & Notation & Original Notation & Previous Works & Short Description \\
        \midrule
        General & & & & \\
        \midrule
        & $C_a$ & $n$ & VOO \cite{kim2020} & Number of action samples \\
        & $C_s$ & $C$ & Sparse Sampling \cite{Kearns2002} & Number of state samples \\
        & & & POWSS \cite{lim2020} & Number of state/observation samples\\
        & $D$ & $H$ & VOO \cite{kim2020} & Horizon of the problem \\
        & & & Sparse Sampling \cite{Kearns2002} & \\
        & $d$ & $h$ & VOO \cite{kim2020} & Given depth of the tree \\
        & $d$ & $h, n$ & Sparse Sampling \cite{Kearns2002} & Given depth of the tree/estimator \\
        \midrule
        VOWSS & & & & \\
        \midrule
        & $\mathcal{R}_{C_a}$ & $\mathcal{R}_n$ & VOO \cite[Theorem~1]{kim2020} & VOO agent regret bound\\
        & $\epsilon$ & $\frac{\lambda}{1-\gamma}$ & POWSS \cite[Theorem~2]{lim2020} & POWSS concentration bound\\
        & $p$ & $\delta$ & POWSS \cite[Theorem~2]{lim2020} & POWSS bound tail probability \\
        \midrule
        VOSS & & & & \\
        \midrule
        & $\mathcal{R}_{C_a}$ & $\mathcal{R}_n$ & VOO \cite[Theorem~1]{kim2020} & VOO agent regret bound\\
        & $\epsilon$ & $\lambda$ & Sparse Sampling \cite[Lemma~3]{Kearns2002} & Chernoff concentration bound\\
        & $p$ & $e^{-\lambda^2C/V_{\max}^2}$ & Sparse Sampling \cite[Lemma~3]{Kearns2002} & Chernoff bound tail probability \\
        \bottomrule
    \end{tabular}
\end{table}

Similarly, Table \ref{tab:concepts} connects the concepts we use to the analogous concepts used by \cite{Kearns2002, kim2020, lim2020}.

\begin{table}[H]
    \centering
    \small
    \caption{Summary of analogous concepts.}
    \label{tab:concepts}
    \begin{tabular}{lllll}
        \toprule
        Proof & Notation & Original Notation & Previous Works & Short Description \\
        \midrule
        VOWSS & & & & \\
        \midrule
        & $\eta_{C_a}(d)$ & $\eta(h)$ & VOO \cite[Theorem~2]{kim2020} & Value function regret bound\\
        & $\alpha_{C_s}(d)$ & $\alpha_d$ & POWSS \cite[Lemma~2]{lim2020} & POWSS recursive bound\\
        \midrule
        VOSS & & & & \\
        \midrule
        & $\eta_{C_a}(d)$ & $\eta(h)$ & VOO \cite[Theorem~2]{kim2020} & Value function regret bound\\
        & $\alpha_{C_s}(d)$ & $\alpha_n$ & Sparse Sampling \cite[Lemma~4]{Kearns2002} & Sparse sampling recursive bound\\
        \bottomrule
    \end{tabular}
\end{table}

\newpage
\subsubsection{VOWSS Conditions}
Before we prove the VOWSS inequality theorem, we present the necessary conditions for this analysis. The conditions required for the proof is the union of regularity conditions required for POWSS and VOO. The POWSS conditions are the following, with appropriate adaptations to our formalism:
\begin{enumerate}[(i)]
    \item (Continuous spaces) $S, A, O$ are continuous spaces. \label{req:space}
    \item (Bounded R\'enyi divergence) For any observation sequence $\{o_n\}_j$, the densities $\mathcal{Z},\mathcal{T},b_0$ are chosen such that the R\'enyi divergence of the target distribution $\mathcal{P}^d$ and sampling distribution $\mathcal{Q}^d$ is bounded above by $d_{\infty}^{\max} < +\infty$ a.s. for all $d = 0,\cdots,D-1$: 
      $$d_{\infty}(\mathcal{P}^d||\mathcal{Q}^d) = \text{ess sup}_{x \sim \mathcal{Q}^d}w_{\mathcal{P}^d/\mathcal{Q}^d}(x) \leq d_{\infty}^{\max}$$
      \label{req:Renyi}
    \item (Bounded reward function) The reward function $R$ is Borel and bounded by a finite constant $||R||_{\infty} \leq R_{\max} < +\infty$ a.s., and $V_{\max} \equiv \frac{R_{\max}}{1-\gamma}<+\infty$. \label{req:r}
    \item (Generative model) We can sample from the generating function $G$ and evaluate the observation probability density $\mathcal{Z}$. \label{req:generate}
    \item (Finite horizon) The POMDP terminates after $D < \infty$ steps. \label{req:finite}
\end{enumerate}
In our context, the target distribution $\mathcal{P}^d$ corresponds to the conditional observation density of the state trajectory, $\mathcal{Q}^d$ the marginal state trajectory density, and $w_{\mathcal{P}^d/\mathcal{Q}^d}$ the importance weights:
\begin{align}\label{eq:pqdef}
  &\mathcal{P}^d = \mathcal{P}_{\{o_n\}_j}^d(\{s_n\}_i) = \frac{(\mathcal{Z}_{1:d}^{i,j}) (\mathcal{T}_{1:d}^i) b_{0}^i}{\int_{S^{d+1}} (\mathcal{Z}_{1:d}^{j}) (\mathcal{T}_{1:d}) b_{0} ds_{0:d}}, \\
  &\mathcal{Q}^d = \mathcal{Q}^d(\{s_n\}_i) = (\mathcal{T}_{1:d}^i) b_{0}^i ,\\
  &w_{\mathcal{P}^d/\mathcal{Q}^d}(\{s_n\}_i) = \frac{(\mathcal{Z}_{1:d}^{i,j}) }{\int_{S^{d+1}} (\mathcal{Z}_{1:d}^{j}) (\mathcal{T}_{1:d}) b_{0} ds_{0:d}} .
\end{align}
Essentially, condition (\ref{req:Renyi}) means that the conditional observation density cannot be much larger than the marginal density for any given state trajectory. 
With the above conditions, the state sampling width $C_s$ is chosen such that the following holds:
\begin{align}
    p &\geq 3C_a(3C_a\cdot C_s)^D\exp(-C_s\cdot t_{\max}^2),\\
    p &= \frac{\epsilon}{(1-\gamma) V_{\max}D},\; t_{\max}(\epsilon,C_s) = \frac{\epsilon(1-\gamma)}{3V_{\max}d_{\infty}^{\max}} - \frac{1}{\sqrt{C_s}}.
\end{align}
This ensures the POWSS type bound $|Q_d^\star(b,a) - \hat{Q}_{d}^\star(\bar{b},a)|\leq \epsilon$ holds with probability at least $1-p$ for each $b,a$ at depth $d.$

The VOO conditions are the following, with appropriate adaptations to our formalism. Note that with our notation, VOO aims to optimize a function $Q(a)$ over some space $A$:
\begin{enumerate}[(i)]
    \item (Translation-invariant semi-metric) $D: A\times A \to \mathbb{R}^+$ is such that $\forall x,y,z \in A,\; D(x,y) = D(y,x),\; D(x,y)=0$ iff $x=y,$ and $D(x+z,y+z) = D(x,y)$.
    \item (Local smoothness of $Q$) There exists at least one global optimum $a^* \in A$ of $Q$ such that $\forall a\in A, Q(a^*)-Q(a)\leq L\cdot D(a,a^*)$ for some $L>0$.
    \item (Shrinkage ratio of the Voronoi cells) Consider any point $a'$ inside the Voronoi cell $C$ generated by the point $a_0$, and denote $d_0 = D(a',a_0).$ If we randomly sample a point $a_1$ from $C$, we have $\mathbb{E}[\min(d_0,D(a',a_1))] \leq \lambda d_0$ for $\lambda \in (0,1)$.
    \item (Well-shaped Voronoi cells) There exists $\eta >0$ such that for any Voronoi cell generated by $a$ with expected diameter $d_0$ contains a ball of radius $\eta d_0$ centered at $a$.
    \item (Local symmetry near optimum) The set of global optima $A_*$ consists of finite number of disjoint and connected components $\{A_*^{(l)}\}_{l=1}^k, k < \infty$. For each component, there exists an open ball $B_{\nu_l}(a_*^{(l)})$ for some $a_*^{(l)} \in A_*^{(l)}$ such that $D(a,a_*^{(l)}) \leq D(a',a_*^{(l)})$ implies $Q(a) \geq Q(a')$ for any $a,a' \in B_{\nu_l}(a_*^{(l)})$.
\end{enumerate}

Here, we define $\bar{\mu}_B(r) = \mu(B_r(\cdot))/\mu(A)$, with $\mu$ the Borel measure on $A$, $\delta_{\max}$ the largest distance between two points in $A$, and $\nu_{\min} = \min_{l \in [k]} \nu_l$.
Then, the action sampling width $C_a$ is chosen such that the following holds:
\begin{align}
    \omega &\geq \frac{1- \lambda^{1/k}}{\bar{\mu}_B(\nu_{\min}) + 1 - \bar{\mu}_B(\eta\cdot\lambda \nu_{\min})},\; C_a \geq \max_{d =0,\cdots, D-1} \lrs{\log \lrp{\frac{\eta_{C_a}(d) - \gamma \eta_{C_a}(d+1)}{2L\delta_{\max}C_{\max}}} \cdot \min (G_{\lambda,\omega}, K_{\nu,\omega,\lambda})}.
\end{align}
Here, $C_{\max}, G_{\lambda,\omega}, K_{\nu,\omega,\lambda}$ are problem specific quantities/functions that are explicitly defined in \cite{kim2020}.
Satisfying these constraints for a decreasing sequence $\{\eta_{C_a}(d)\}$ will allow us to use the VOO type bound $V_d^\star(b) - \hat{V}_d^\star(b) \leq \eta_{C_a}(d)$ that holds in expectation.

\subsubsection{Proof of VOWSS Inequality}
\begin{theorem}[VOWSS Inequality] \label{thm:app-vowss}
    Suppose we choose the action sampling width $C_a$ and state sampling width $C_s$ such that under the union of regularity conditions specified by \cite{lim2020} and \cite{kim2020}, the intermediate POWSS bounds and VOO bounds in Lemma \ref{lemm:app-vowss} are satisfied at every depth of the tree. Then, the following bounds for the VOWSS estimator $\hat{V}_{\textsc{VOWSS},d}(\bar{b})$ hold for all $d\in[0,D-1]$ in expectation:
    \begin{align}
        \lra{V_d^\star(b) - \hat{V}_{\textsc{VOWSS},d}(\bar{b})} &\leq \eta_{C_a} + \alpha_{C_s}.
    \end{align}
\end{theorem}

In order to prove Theorem \ref{thm:app-vowss}, we first prove an intermediate lemma which will allow us to obtain the bound through triangle inequality. 
We introduce and prove the following lemma first. 
All of the following calculations are done in expectation.
We also denote $\bar{b}$ for a particle representation of belief $b$ that POWSS and VOWSS take as an argument.

\begin{lemma}[VOWSS Intermediate Inequality] \label{lemm:app-vowss}
  Suppose with our notation, the POWSS estimators at all depths $d$ are within $\epsilon$ of their mean values with probability $1-p$, and the VOO agents have regret bounds of $\mathcal{R}_{C_a}$.
  The following inequalities hold for all $d\in[0,D-1]$ in expectation:
  \begin{align}
    \lra{V_d^\star(b) - \hat{V}_d^{C_a}(b)} &\leq \eta_{C_a}(d),\\
    \lra{\hat{V}_d^{C_a}(b) - \hat{V}_{\textsc{VOWSS},d}(\bar{b})} &\leq \alpha_{C_s}(d).
  \end{align}
  $\eta_{C_a}(d)$ and $\alpha_{C_s}(d)$ are sequences that satisfy the following properties:
  \begin{align}
    \eta_{C_a}(d) &\geq \gamma \cdot \eta_{C_a}(d+1) + \mathcal{R}_{C_a},\; \eta_{C_a}(D) = 0,\\
    \eta_{C_a} &\equiv \max_{d=0,\cdots,D-1} \eta_{C_a}(d)< + \infty ,\\
    \alpha_{C_s}(d) &\equiv (1+\gamma)(\epsilon + 2p\cdot V_{\max}) + \gamma (\alpha_{C_s}(d+1) + 2p\cdot V_{\max}),\\ 
    \alpha_{C_s}(D-1) &= \epsilon + 2p\cdot V_{\max},\\
    \alpha_{C_s} &\equiv \max_{d=0,\cdots,D-1} \alpha_{C_s}(d)< + \infty.
  \end{align}
\end{lemma}
\proof This proof proceeds through induction by assuming that the bounds hold for all depths from $d+1$ to $D-1$, and then proving they also hold for depth $d$ (induction hypothesis is omitted as the bounds trivially hold as per the definitions of VOOT and POWSS). We divide the main inequality into VOO bound and POWSS bound by introducing an intermediate term $\hat{V}_d^{C_a}(b)$:
\begin{align}
  \lra{V_d^\star(b) - \hat{V}_{\textsc{VOWSS},d}(b)} &\leq \undb{\lra{V_d^\star(b) - \hat{V}_d^{C_a}(b)}}{VOO bound} + \undb{\lra{\hat{V}_d^{C_a}(b) - \hat{V}_{\textsc{VOWSS},d}(\bar{b})}}{POWSS bound}
\end{align}
Essentially, we have two main layers of inequality, caused by the stochastic nature of VOO action selection, and the uncertainties in state transition and observation. 
We will first analyze the VOO bound, then the POWSS bound.

\paragraph{Step 1: VOO Bound}
The VOO bound can once again be further decomposed into the following terms as \cite{kim2020} do in their work:
\begin{align}
  \lra{V_d^\star(b) - \hat{V}_d^{C_a}(b)}&\leq \undb{\lra{V_d^\star(b) - \hat{V}_d(b)}}{VOO Recursive bound} + \undb{\lra{\hat{V}_d(b) - \hat{V}_d^{C_a}(b)}}{VOO Regret bound}
\end{align}
Here, the intermediate random variables are defined in the following manner:
\begin{align}
 V_d^\star(b)&\equiv \max_{a\in A}Q^\star_d(b,a) = \max_{a\in A}\lrc{R(b,a) + \gamma \E[V^\star_{d+1}(bao)|b]}\\
  \hat{V}_d(b) &\equiv \max_{a\in A}\hat{Q}_d(b,a)= \max_{a\in A}\lrc{R(b,a) + \gamma \E[\hat{V}_{d+1}^{C_a}(bao)|b]}\\
  \hat{V}_d^{C_a}(b) &\equiv \max_{a\in VOO(A,C_a)}\hat{Q}_d(b,a)= \max_{a\in VOO(A,C_a)}\lrc{R(b,a) + \gamma \E[\hat{V}_{d+1}^{C_a}(bao)|b]}
\end{align}
As a notation, $a\in VOO(A,C_a)$ indicates that the actions $a$ are chosen sequentially through the VOO algorithm over the action space $A$ for $C_a$ iterations of VOO.
Here, we compare the two quantities with a reference action $a^\star = \argmax Q^\star_d(b,a)$, which results in a looser bound but allows us to directly compare the quantities inside the max operations.
We closely follow the calculations from the proof of Lemma 5 in \cite{kim2020}: 
\begin{align}
  \undb{\lra{V_d^\star(b) - \hat{V}_d(b)}}{VOO Recursive bound} &= \lra{\max_{a\in A}\lrc{R(b,a) + \gamma \E[V^\star_{d+1}(bao)|b]} - \max_{a\in A}\lrc{R(b,a) + \gamma \E[\hat{V}_{d+1}^{C_a}(bao)|b]}}\\
  &\leq \lra{R(b,a^\star ) + \gamma \E[V^\star_{d+1}(ba^\star o)|b] - R(b,a^\star ) + \gamma \E[\hat{V}_{d+1}^{C_a}(ba^\star o)|b]} \tag{$a^\star  = \argmax Q^\star_d(b,a)$}\\
  &\leq \gamma \E\lrs{|V^\star_{d+1}(bao) - \hat{V}_{d+1}^{C_a}(bao)| \given b} \\
  &\leq \gamma \cdot \eta_{C_a}(d+1)\\
  \undb{\lra{\hat{V}_d(b) - \hat{V}_d^{C_a}(b)}}{VOO Regret bound} &\leq \mathcal{R}_{C_a}.
\end{align}
In the VOO Regret bound, the difference cannot be less than zero since the global maximum is at least as big as the VOO maximum, so the absolute value disappears and we get the regret of VOO.
Thus, with our choice of $C_a$ that is designed to satisfy the recurrence relation:
\begin{align}
  \lra{V_d^\star(b) - \hat{V}_d^{C_a}(b)}&\leq \undb{\lra{V_d^\star(b) - \hat{V}_d(b)}}{VOO Recursive bound} + \undb{\lra{\hat{V}_d(b) - \hat{V}_d^{C_a}(b)}}{VOO Regret bound}\\
  &\leq \gamma \cdot \eta_{C_a}(d+1) + \mathcal{R}_{C_a} \leq \eta_{C_a}(d).
\end{align}

\paragraph{Step 2: POWSS Bound}
The POWSS bound can also be further decomposed into the following terms:
\begin{align}
  \lra{\hat{V}_d^{C_a}(b) - \hat{V}_{\textsc{VOWSS},d}(b)} &\leq \undb{\lra{\hat{V}_d^{C_a}(b) - \tilde{V}_{\textsc{VOWSS},d}(b)}}{POWSS Concentration bound} + \undb{\lra{\tilde{V}_{\textsc{VOWSS},d}(b) - \hat{V}_{\textsc{VOWSS},d}(b)}}{POWSS Recursive bound}.
\end{align}
Here, the extra intermediate random variable and the VOWSS estimator are defined in the following manner:
\begin{align}
  \tilde{V}_{\textsc{VOWSS},d}(b) &\equiv \max_{a\in VOO(A,C_a)}\tilde{Q}_{\textsc{VOWSS},d}(b,a)= \max_{a\in VOO(A,C_a)}\lrc{\frac{\sum_{i=1}^{C_s}w_{d,i}(r_{d,i} + \gamma \hat{V}_{d+1}^{C_a}(bao_i))}{\sum_{i=1}^{C_s}w_{d,i}}},\\
  \hat{V}_{\textsc{VOWSS},d}(\bar{b}) &\equiv \max_{a\in VOO(A,C_a)}\hat{Q}_{\textsc{VOWSS},d}(\bar{b},a)= \max_{a\in VOO(A,C_a)}\lrc{\frac{\sum_{i=1}^{C_s}w_{d,i}(r_{d,i} + \gamma \hat{V}_{\textsc{VOWSS},d+1}(\overline{bao_i}))}{\sum_{i=1}^{C_s}w_{d,i}}}.
\end{align}
We compare the two quantities by picking a reference action to directly compare the quantities inside the max operations. 

For the concentration bound term:
\begin{align}
  \undb{\lra{\hat{V}_d^{C_a}(b) - \tilde{V}_{\textsc{VOWSS},d}(b)}}{POWSS Concentration bound} &\leq \ilim{}{}\max_{a\in VOO(A,C_a)}\lrc{R(b,a) + \gamma \E[\hat{V}_{d+1}^{C_a}(bao)|b]} \\
  &\;\;\;\;\;\;\;\;- \max_{a\in VOO(A,C_a)}\lrc{\frac{\sum_{i=1}^{C_s}w_{d,i}(r_{d,i} + \gamma \hat{V}_{d+1}^{C_a}(bao_i))}{\sum_{i=1}^{C_s}w_{d,i}}} \ilim{}{} \\
  &\leq \lra{R(b,a^*) + \gamma \E[\hat{V}_{d+1}^{C_a}(ba^*o)|b] - \frac{\sum_{i=1}^{C_s}w_{d,i}(r_{d,i} + \gamma \hat{V}_{d+1}^{C_a}(ba^*o_i))}{\sum_{i=1}^{C_s}w_{d,i}}}. \tag{$a^* = \argmax_{VOO} \hat{Q}_d(b, a)$}
\end{align}
As a small note, picking the maximizing action with respect to $\hat{Q}_d$ does not guarantee that we have bounded the term in this case.
We also need to consider the case of picking the maximizing action with respect to $\tilde{Q}_{\textsc{VOWSS},d}$, since $\tilde{Q}_{\textsc{VOWSS},d}$ could achieve larger values than $\hat{Q}_d$ due to sparse sampling of states.
However, our overall result does not change, since choosing the other maximizing action will only result in a change in sign within the absolute value, which we can still bound with the recursive bound.
Thus, in the following calculations whenever we pick the reference action for difference in quantities that do not have strict magnitude hierarchy, we do not  consider this possibility as the terms can still be bounded with their respective recursive bounds by choosing an appropriate reference action.

Decomposing the quantity into the reward difference and the next-step value difference:
\begin{align}
  \undb{\lra{\hat{V}_d^{C_a}(b) - \tilde{V}_{\textsc{VOWSS},d}(b)}}{POWSS Concentration bound} &\leq \undb{\lra{R(b,a^*) - \frac{\sum_{i=1}^{C_s}w_{d,i}r_{d,i}}{\sum_{i=1}^{C_s}w_{d,i}}}}{Reward difference}+ \gamma \undb{\lra{\E[\hat{V}_{d+1}^{C_a}(ba^*o)|b] - \frac{\sum_{i=1}^{C_s}w_{d,i}\hat{V}_{d+1}^{C_a}(ba^*o_i)}{\sum_{i=1}^{C_s}w_{d,i}}}}{Value difference}.
\end{align}
For the reward difference, we can crudely upper bound it by the POWSS $Q$-value estimate concentration bound in Theorem 2 of \cite{lim2020}, since this is effectively the same structure as the leaf node estimate. 
In the POWSS concentration bound in Theorem 2, the difference has an upper bound of $\lambda/(1-\gamma)$, which we will define as $\epsilon$, and we denote the corresponding probability $\delta$ as $p$.
This bounds the quantity by $(\epsilon + 2p\cdot V_{\max})$ using the expectation version of the POWSS concentration bound.
\begin{align}
  \undb{\lra{R(b,a^*) - \frac{\sum_{i=1}^{C_s}w_{d,i}r_{d,i}}{\sum_{i=1}^{C_s}w_{d,i}}}}{Reward difference} &\leq \epsilon + 2p\cdot V_{\max}.
\end{align}
We could instead use Lemma 1 of \cite{lim2020}, the leaf node estimate concentration bound, but we use the more general Theorem 2.
This allows us to consistently use the same theorem throughout this analysis and effectively combine terms.
Similarly, while the worst case bound is $2R_{\max}$ since we are taking the difference of two reward functions, we crudely upper bound that with $2V_{\max}$ for algebraic convenience of combining it with the value difference term.
\begin{align}
  \undb{\lra{\E[\hat{V}_{d+1}^{C_a}(ba^*o)|b] - \frac{\sum_{i=1}^{C_s}w_{d,i}\hat{V}_{d+1}^{C_a}(ba^*o_i)}{\sum_{i=1}^{C_s}w_{d,i}}}}{Value difference} \leq \epsilon + 2p\cdot V_{\max}.
\end{align}
On the other hand, for the value difference, the POWSS concentration bound also turns out to be an upper bound, but with more sophisticated calculations. 
During this part of the proof, we will refer heavily back to the continued proof of Lemma 2 in the Appendix C of \cite{lim2020} and give a general overview of how the steps apply here. 

\paragraph{Step 2-i: Value Difference}
While Lemma 2 in \cite{lim2020} is calculated with respect to the theoretically optimal value function $V^*_{d+1}$, the calculation steps themselves and the theorems and lemmas used there can apply exactly the same way for $\hat{V}_{d+1}^{C_a}$.
We will briefly illustrate how the steps are parallel to the proof of Lemma 2 in the Appendix C of \cite{lim2020}.

The value difference corresponds to the difference between the expected value of $\hat{V}_{d+1}^{C_a}$ and the self-normalized importance sampling estimator of the expected value. 
This value difference specifically corresponds to the first two error terms in the continued proof of Lemma 2 in the Appendix C of \cite{lim2020}. 
The decomposition looks like the following:
\begin{align}
    &\lra{\E[\hat{V}_{d+1}^{C_a}(ba^*o)|b] - \frac{\sum_{i=1}^{C_s}w_{d,i}\hat{V}_{d+1}^{C_a}(ba^*o_i)}{\sum_{i=1}^{C_s}w_{d,i}}}\\
    &\leq \undb{\lra{\E[\hat{V}_{d+1}^{C_a}(ba^*o)|b] - \frac{\sum_{i=1}^{C_s}w_{d,i}\hat{\mathbf{V}}_{d+1}^{C_a}(s_{d,i}, b, a^*)}{\sum_{i=1}^{C_s}w_{d,i}}}}{Importance sampling error} + \undb{\lra{\frac{\sum_{i=1}^{C_s}w_{d,i}(\hat{\mathbf{V}}_{d+1}^{C_a}(s_{d,i}, b, a^*) - \hat{V}_{d+1}^{C_a}(ba^*o_i))}{\sum_{i=1}^{C_s}w_{d,i}}}}{MC next-step integral approximation error}.
\end{align}
The next-step marginal integral $\hat{\mathbf{V}}_{d+1}^{C_a}(s_{d,i}, b, a^*)$ is defined as
\begin{align}
    \hat{\mathbf{V}}_{d+1}^{C_a}(s_{d,i}, b, a^*) &\equiv \int_S\int_O \hat{V}_{d+1}^{C_a}(ba^*o)\mathcal{Z}(o|a^*,s_{d+1})\mathcal{T}(s_{d+1}|s_{d,i},a^*)ds_{d+1}do.
\end{align}
With this analogous definition, the self-normalized estimator identities in \cite{lim2020} can be exactly applied to this setting once again, now instead for the function $\hat{V}_d^{C_a}$ because the algebraic steps taken in the proof should hold for our $V$ function estimator as well.
Intuitively, the next-step marginal integral is defined to be the marginal random variable for $\hat{V}_d^{C_a}$, instead of the optimal $V$ function as it is done in the works of \cite{lim2020}.

First, following \cite{lim2020}, we define the following notation for products of transition and observation densities:
\begin{align}
  \mathcal{T}_{1:d}^i &\equiv \prod_{n=1}^d \mathcal{T}(s_{n,i}|s_{n-1,i},a_n), \\
  \mathcal{Z}_{1:d}^{i,j} &\equiv \prod_{n=1}^d \mathcal{Z}(o_{n,j}|a_{n},s_{n,i}).
\end{align}
Specifically, $i$ denotes the index of the state sample, and $j$ denotes the index of the observation sample.
Absence of any of the indices $i$ or $j$ means that the state trajectory $\{s_n\}$ or the observation history $\{o_n\}$ appear as regular variables, mostly for the purposes of integration.

\paragraph{Step 2-ii: Importance Sampling Error}
For the importance sampling error, note that for a belief $b$ at depth $d,$
\begin{align}
    \E[\hat{V}_{d+1}^{C_a}(ba^*o)|b] &= \int_S \int_S \int_O \hat{V}_{d+1}^{C_a}(ba^*o)(\mathcal{Z}_{d+1}) (\mathcal{T}_{d,d+1})b \cdot ds_{d:d+1}do\\
    &= \int_S \hat{\mathbf{V}}_{d+1}^{C_a}(s_{d,i}, b, a^*)b\cdot ds_{d}\\
    &= \frac{\int_{S^{d}} \hat{\mathbf{V}}_{d+1}^{C_a}(s_{d,i}, b, a^*) (\mathcal{Z}_{1:d}) (\mathcal{T}_{1:d}) b_{0} ds_{0:d}}{\int_{S^{d}} (\mathcal{Z}_{1:d}) (\mathcal{T}_{1:d}) b_{0} ds_{0:d}}.
\end{align}
Consequently, the weighted average of the next-step marginal integral $\hat{\mathbf{V}}_{d+1}^{C_a}(s_{d,i}, b, a^*)$ is a self-normalized importance sampling estimator of $\hat{V}_{d+1}^{C_a}(ba^*o)$ given $b$, and we can apply the augmented self-normalized estimator concentration bound in the same way as \cite{lim2020} to get the concentration bound of $\lambda/3$.

\paragraph{Step 2-iii: Monte Carlo Next-Step Integral Approximation Error}
For the MC next-step integral approximation error, generating estimates of $\hat{V}_{d+1}^{C_a}(ba^*o_i)$ for a given $(s_{d,i}, b, a^*)$ also results in an unbiased Monte Carlo estimator of the next-step marginal integral, and the following difference is mean zero conditioned on $(s_{d,i}, b, a^*)$:
\begin{align}
  \Delta_{d+1}(s_{d,i},b,a^*) &\equiv \hat{\mathbf{V}}_{d+1}^{C_a}(s_{d,i}, b, a^*) - \hat{V}_{d+1}^{C_a}(ba^*o_i).
\end{align}
Thus, the same calculation steps hold. 
Note that in the works of \cite{lim2020}, the MC next-step integral approximation error is further crudely bound by $\frac{2}{3\gamma}\lambda$, when the actual bounds also hold for $\frac{2}{3}\lambda$, for the convenience of being able to combine the $\gamma$ multiplied terms in the main proof of Lemma 2.
Thus, we can bound the MC next-step integral approximation error with the stricter bound $\frac{2}{3}\lambda$.

In our work, we used the variable $\epsilon$ to denote the POWSS concentration bound, which corresponds to $\epsilon = \lambda/(1-\gamma) \geq \lambda$.
Since the sum of importance sampling error and MC next-step integral approximation error are bounded by $(1/3 + 2/3)\lambda = \lambda$, this can also further be crudely bounded by the POWSS concentration inequality with the upper bound $\epsilon$.

We have now obtained bounds for the value difference term using the expectation version of the POWSS concentration inequality where the extreme probability event is once again bounded with the term $2p\cdot V_{\max}$:
\begin{align}
    \lra{\E[\hat{V}_{d+1}^{C_a}(ba^*o)|b] - \frac{\sum_{i=1}^{C_s}w_{d,i}\hat{V}_{d+1}^{C_a}(ba^*o_i)}{\sum_{i=1}^{C_s}w_{d,i}}} \leq \epsilon + 2p\cdot V_{\max}.
\end{align}

Finally, we can bound the POWSS Concentration bound term:
\begin{align}
  \undb{\lra{\hat{V}_d^{C_a}(b) - \tilde{V}_{\textsc{VOWSS},d}(b)}}{POWSS Concentration bound} &\leq (\epsilon + 2p\cdot V_{\max}) + \gamma(\epsilon + 2p\cdot V_{\max})
  = (1+\gamma)(\epsilon + 2p\cdot V_{\max}).
\end{align}

For the POWSS Recursive bound term, we simply apply the inductive hypothesis for step $d+1$:
\begin{align}
  \undb{\lra{\tilde{V}_{\textsc{VOWSS},d}(b) - \hat{V}_{\textsc{VOWSS},d}(\bar{b})}}{POWSS Recursive bound} &\leq \ilim{}{}\max_{a\in VOO(A,C_a)}\lrc{\frac{\sum_{i=1}^{C_s}w_{d,i}(r_{d,i} + \gamma \hat{V}_{d+1}^{C_a}(bao_i))}{\sum_{i=1}^{C_s}w_{d,i}}} \\
  &\;\;\;\;\;\;\;\;- \max_{a\in VOO(A,C_a)}\lrc{\frac{\sum_{i=1}^{C_s}w_{d,i}(r_{d,i} + \gamma \hat{V}_{\textsc{VOWSS},d+1}(\overline{bao_i}))}{\sum_{i=1}^{C_s}w_{d,i}}} \ilim{}{} \\
  &\leq \lra{\gamma\frac{\sum_{i=1}^{C_s}w_{d,i}(\hat{V}_{d+1}^{C_a}(b\tilde{a}o_i)-\hat{V}_{\textsc{VOWSS},d+1}(\overline{b\tilde{a}o_i}))}{\sum_{i=1}^{C_s}w_{d,i}}} \tag{$\tilde{a} = \argmax_{VOO} \tilde{Q}_{VOWSS,d}(b, a)$}\\
  &\leq \gamma\frac{\sum_{i=1}^{C_s}w_{d,i}\lra{\hat{V}_{d+1}^{C_a}(b\tilde{a}o_i)-\hat{V}_{\textsc{VOWSS},d+1}(\overline{b\tilde{a}o_i})}}{\sum_{i=1}^{C_s}w_{d,i}}\\
  &\leq \gamma \cdot \alpha_{C_s}(d+1).
\end{align}
Putting the POWSS components together, we prove the POWSS bound by induction:
\begin{align}
  \lra{\hat{V}_d^{C_a}(b) - \hat{V}_{\textsc{VOWSS},d}(\bar{b})} &\leq \undb{\lra{\hat{V}_d^{C_a}(b) - \tilde{V}_{\textsc{VOWSS},d}(b)}}{POWSS Concentration bound} + \undb{\lra{\tilde{V}_{\textsc{VOWSS},d}(b) - \hat{V}_{\textsc{VOWSS},d}(\bar{b})}}{POWSS Recursive bound}\\
  &\leq (1+\gamma)(\epsilon + 2p\cdot V_{\max}) + \gamma \cdot \alpha_{C_s}(d+1) = \alpha_{C_s}(d). \done
\end{align}

\textit{Proof for Theorem \ref{thm:app-vowss}}. Finally, we prove the main theorem by combining the two terms:
\begin{align}
  \lra{V_d^\star(b) - \hat{V}_{\textsc{VOWSS},d}(\bar{b})} &\leq \undb{\lra{V_d^\star(b) - \hat{V}_d^{C_a}(b)}}{VOO bound} + \undb{\lra{\hat{V}_d^{C_a}(b) - \hat{V}_{\textsc{VOWSS},d}(b)}}{POWSS bound} \leq \eta_{C_a}(d) + \alpha_{C_s}(d) \leq \eta_{C_a} + \alpha_{C_s}. \done
\end{align}

Here, the $\eta_{C_a}(d)$ corresponds to the VOO bound in \cite{kim2020}.
Thus, we can always find a corresponding $C_a$ for an arbitrary decreasing sequence of $\eta_{C_a}(d)$ which satisfies the properties in our lemma, and this sequence can be made closer to 0 for each depth by choosing bigger $C_a$ values.

On the other hand, $\alpha_{C_s}(d)$ is comprised of POWSS bound components in \cite{lim2020}, which decreases in both $p$ and $\epsilon$ with more samples $C_s$. Thus, it can also be made to decrease to 0 as we increase $C_s$.

\newpage
\subsection{Experiment Hyperparameters}
Hyperparameters were taken from the references if they were given, and otherwise the set of hyperparmaters for a system was obtained by first training POMCPOW, and then training VOMCPOW and BOMCP centered around POMCPOW hyperparameters as initial estimates. 
This usually augmented the performances of VOMCPOW and BOMCP compared to directly borrowing the POMCPOW hyperparameters. 
For BOMCP, buffer and $k$ were fixed at 100 and 5, respectively, as it was done in the original paper by \cite{mern2020}.
Table \ref{tab:hyper} shows the final hyperparameters used in the experiments.
Specifically for the lunar lander problem, we have chosen the dynamics time step $dt=0.4$, which gave us the most consistent performance of BOMCP with 100 queries over 1000 iterations that is in line with the results of \cite{mern2020}.

We give a brief intuitive explanation for each of the parameters utilized.
$c$ is the critical factor or the Upper Confidence Bound exploration parameter, which governs how much we should explore among the action samples we have collected.
Since VPW needs to balance local and global search, on top of estimating relatively faithful $Q$-values, it usually worked better to set the $c$ on the equal magnitude/lower than the $c$ value for POMCPOW, usually around 10-100.
$k_a$ is the constant factor of the action widening parameter, which sets the overall width of the action widening.
$\alpha_a$ is the exponential factor of the action widening parameter, which determines how adaptive we want to progressively widen the action width.
$k_o,\alpha_o$ are the state/observation widening parameters, which function similarly to the action widening parameters.
$\omega$ is the VOO exploration probability, which governs how much we should explore in the action space.
Since in global search, we need to see enough samples to explore the action space sufficiently, it usually worked better to set $\omega$ to be relatively high around 0.7-0.9.
Lastly, $\Sigma$ is the Gaussian covariance matrix for VOO rejection sampling.

The only hyperparameter that was manually picked is the Gaussian covariance matrix for VOO rejection sampling. 
When picking this covariance matrix, we usually found that it was most effective to use a diagonal matrix with entries that are around 10 to 20 times less than the maximum action space bounds for each dimension. 
In theory, the diagonal covariance matrix can also be fitted via CEM, but we chose to handpick these values after some inspection in order to reduce the number of hyperparameters that needed to be fit.

To limit the number of rejection sampling iterations, we set the maximum number of rejection sampling iterations to 20 and automatically choose the closest action we have sampled when we reach 20 iterations.
We also manually set automatic sample acceptance regions that was usually on the order of a tenth of the sampling radius, which we found was not strictly necessary in conjunction with the sampling iteration limit.

The open source code is available at \texttt{github.com/michael-lim/VOOTreeSearch.jl}.

\begin{table}[h]
    \centering
    \small
    \caption{Summary of hyperparameters used in experiments for POMCPOW, VOMCPOW, BOMCP, and VOWSS.}
    \label{tab:hyper}
    \begin{tabular}{lcccccccccccc}
        \toprule
        & $c$ & $k_a$ & $\alpha_a$ & $k_o$ & $\alpha_o$ & $C_s$ & $C_a$ & $\omega$ & $\Sigma=$ diag($\sigma^2$) & $l$ & $\lambda$ & $\gamma_a$\\ 
        \midrule
        LQG (Depth 3) & & & & & & & & & & &\\
        \midrule
        POMCPOW & 65.0& 30.0& $\frac{1}{2.5}$& 30.0&$\frac{1}{4}$ & & & & & &\\
        VOMCPOW & 60.0& 25.0& $\frac{1}{5.5}$& 25.0& $\frac{1}{2.5}$& & & 0.8& [0.5, 0.5]& & &\\
        BOMCP & 135.0& 30.0& $\frac{1}{4}$& 20.0& $\frac{1}{4}$& & & & & $\log(15)$ & 0.4\\
        VOWSS & & & & & & 10& 200& 0.8& [0.5, 0.5]& & & 0.4\\
        \midrule
        VDP Tag (Depth 10) & & & & & & & & & & &\\
        \midrule
        POMCPOW & 110.0& 30.0& $\frac{1}{30}$& 5.0& $\frac{1}{100}$& & & & & &\\
        VOMCPOW & 85.0& 30.0& $\frac{1}{30}$& 2.5& $\frac{1}{100}$& & & 0.7& [0.1]& &\\
        \midrule
        Lunar Lander (Depth 250) & & & & & & & & & & &\\
        \midrule
        POMCPOW & 10.0& 3.0& $\frac{1}{4}$& 2.0& $\frac{1}{10}$& & & & & &\\
        VOMCPOW & 30.0& 4.0& $\frac{1}{4}$& 1.5& $\frac{1}{5}$& & & 0.9& [0.2, 0.5, 0.05]& &\\
        BOMCP & 10.0& 3.0& $\frac{1}{4}$& 2.0& $\frac{1}{10}$& & & & & $\log(15)$ & 0.5\\
        \bottomrule
    \end{tabular}
\end{table}
        
\clearpage
\subsection{Voronoi Optimistic Sparse Sampling (VOSS) for Stochastic MDPs}

Voronoi Optimistic Sparse Sampling (VOSS) is an application of VPW to the sparse sampling algorithm \cite{Kearns2002} to tackle the stochastic MDP case.
Like VOWSS, it can be defined by an \textsc{EstimateQ} function that estimates the $Q$-value with next-step state samples.
Since VOSS does not have observation uncertainty, it only needs to take the arithmetic average instead of the observation likelihood weighted average.
The \textsc{EstimateQ} function that takes in a state $s$ instead of a belief particle set $\bar{b}$ is outlined in Algorithm \ref{alg:app-voss}.
Note that this is analogous definition to the $Q$-function estimation algorithm in \cite{Kearns2002}.
The \textsc{EstimateV} function should function similarly, except working with $s$ instead of $\bar{b}$.

\begin{algorithm}[H]
\caption{Value estimation algorithm for VOSS} \label{alg:app-voss}
\textbf{Algorithm:} \textsc{EstimateQ}($s, a, d$)\\
\textbf{Input:} State $s$, action $a$, depth $d$.\\
\textbf{Output:} A scalar $\hat{Q}_d^\star(s,a)$ that is an estimate of $Q_d^\star(s,a)$.
\begin{algorithmic}[1]
\State For $ i=1,\cdots,C_s$, generate $s_i',r = G(s,a)$.
\State Return the $Q$-value estimate:
\end{algorithmic}
\begin{align*}
  &\hat{Q}_d^\star(s,a) = r + \gamma \frac{1}{C_s}\sum_{i = 1}^{C_s}\textsc{EstimateV}(s_i', d+1) .
\end{align*}
\end{algorithm}

The conditions required for the proof is the union of regularity conditions required for sparse sampling and VOO. 
The sparse sampling conditions are the subset of POWSS conditions.
Namely, we only require the conditions (\ref{req:space}), (\ref{req:r}), (\ref{req:generate}), only for the state and action spaces.
Since in this proof we simply use the Chernoff bound for the sparse sampling type bound, we just need to make sure that $p \geq \exp(-\epsilon^2C_s/(2V_{\max})^2)$ holds when picking $C_s$.

\begin{theorem}[VOSS Inequality] \label{thm:app-voss}
  Suppose we choose the action sampling width $C_a$ and state sampling width $C_s$ such that under the union of regularity conditions specified by \cite{Kearns2002} and \cite{kim2020}, the intermediate sparse sampling bounds and VOO bounds in Lemma \ref{lemm:app-voss} are satisfied at every depth of the tree. Then, the following bounds for the VOSS estimator $\hat{V}_{\textsc{VOSS},d}(s)$ hold for all $d\in[0,D-1]$ in expectation:
  \begin{align}
    \lra{V_d^\star(s) - \hat{V}_{\textsc{VOSS},d}(s)} &\leq \eta_{C_a} + \alpha_{C_s}. \label{eq:voss}
  \end{align}
\end{theorem}

Similar to proving Theorem \ref{thm:app-vowss}, to prove Theorem \ref{thm:app-voss}, we first prove an intermediate lemma which will allow us to obtain the bound through triangle inequality. 
The proof is easier than that of Theorem \ref{thm:app-vowss}, since we do not explicitly need to deal with the observation uncertainty.
We introduce and prove the following lemma first.
All of the following calculations are done in expectation.

\begin{lemma}[VOSS Intermediate Inequality] \label{lemm:app-voss}
  Suppose with our notation, the sparse sampling estimators at all depths $d$ are within $\epsilon$ of their mean values with probability $1-p$, and the VOO agents have regret bounds of $\mathcal{R}_{C_a}$.
  The following inequalities hold for all $d\in[0,D-1]$ in expectation:
  \begin{align}
    \lra{V_d^\star(s) - \hat{V}_d^{C_a}(s)} &\leq \eta_{C_a}(d), \\
    \lra{\hat{V}_d^{C_a}(s) - \hat{V}_{\textsc{VOSS},d}(s)} &\leq \alpha_{C_s}(d).
  \end{align}
  $\eta_{C_a}(d)$ and $\alpha_{C_s}(d)$ are sequences that satisfy the following properties:
  \begin{align}
      \eta_{C_a}(d) &\geq \gamma \cdot \eta_{C_a}(d+1) + \mathcal{R}_{C_a},\; \eta_{C_a}(D) = 0,\\
      \eta_{C_a} &\equiv \max_{d=0,\cdots,D-1} \eta_{C_a}(d)< + \infty,\\
      \alpha_{C_s}(d) &\equiv \gamma(\alpha_{C_s}(d+1) + \epsilon + 2p\cdot V_{\max}),\; \alpha_{C_s}(D-1) = \epsilon + 2p\cdot V_{\max},\\
      \alpha_{C_s} &\equiv \max_{d=0,\cdots,D-1} \alpha_{C_s}(d)< + \infty.
  \end{align}
\end{lemma}
\proof This proof proceeds through induction by assuming that the bounds hold for all depths from $d+1$ to $D-1$, and then proving they also hold for depth $d$ (induction hypothesis is omitted as the bounds trivially hold as per the definitions of VOOT and sparse sampling). We divide the main inequality into VOO bound and sparse sampling bound (SS bound) by introducing an intermediate term $\hat{V}_d^{C_a}(s)$:
\begin{align}
  \lra{V_d^\star(s) - \hat{V}_{\textsc{VOSS},d}(s)} &\leq \undb{\lra{V_d^\star(s) - \hat{V}_d^{C_a}(s)}}{VOO bound} + \undb{\lra{\hat{V}_d^{C_a}(s) - \hat{V}_{\textsc{VOSS},d}(s)}}{SS bound}.
\end{align}
We now have two main layers of inequality, caused by the stochastic nature of VOO action selection and the uncertainty in state transition. 
We will first analyze the VOO bound, then the SS bound.

\paragraph{Step 1: VOO Bound}
The VOO bound can be further decomposed into the following terms as \cite{kim2020} do in their work:
\begin{align}
  \lra{V_d^\star(s) - \hat{V}_d^{C_a}(s)}&\leq \undb{\lra{V_d^\star(s) - \hat{V}_d(s)}}{VOO Recursive bound} + \undb{\lra{\hat{V}_d(s) - \hat{V}_d^{C_a}(s)}}{VOO Regret bound}.
\end{align}
This step is very close to our previous procedure in VOWSS. Here, the intermediate random variables are defined in the following manner:
\begin{align}
  V_d^\star(s)&\equiv \max_{a\in A}Q^\star_d(s,a) = \max_{a\in A}\lrc{R(s,a) + \gamma \E_{s'\sim T(s,a)}[V^\star_{d+1}(s')]},\\
  \hat{V}_d(s) &\equiv \max_{a\in A}\hat{Q}_d(s,a)= \max_{a\in A}\lrc{R(s,a) + \gamma \E_{s'\sim T(s,a)}[\hat{V}_{d+1}^{C_a}(s')]},\\
  \hat{V}_d^{C_a}(s) &\equiv \max_{a\in VOO(A,C_a)}\hat{Q}_d(s,a)= \max_{a\in VOO(A,C_a)}\lrc{R(s,a) + \gamma \E_{s'\sim T(s,a)}[\hat{V}_{d+1}^{C_a}(s')]}.
\end{align}
Repeating the procedures in Lemma \ref{lemm:app-vowss}, we closely follow the calculations from Lemma 5 in \cite{kim2020}: 
\begin{align}
  \undb{\lra{V_d^\star(s) - \hat{V}_d(s)}}{VOO Recursive bound} &= \lra{\max_{a\in A}\lrc{R(s,a) + \gamma \E_{s'\sim T(s,a)}[V^\star_{d+1}(s')]} - \max_{a\in A}\lrc{R(s,a) + \gamma \E_{s'\sim T(s,a)}[\hat{V}_{d+1}^{C_a}(s')]}}\\
  &\leq \lra{R(s,a^\star) + \gamma \E_{s'\sim T(s,a^\star)}[V^\star_{d+1}(s')] - R(s,a^\star) - \gamma \E_{s'\sim T(s,a^\star)}[\hat{V}_{d+1}^{C_a}(s')]} \tag{$a^\star = \argmax Q^\star_d(s,a)$}\\
  &\leq \gamma \E_{s'\sim T(s,a^\star)}\lra{V^\star_{d+1}(s') - \hat{V}_{d+1}^{C_a}(s')} \\
  &\leq \gamma \cdot \eta_{C_a}(d+1)\\
  \undb{\lra{\hat{V}_d(s) - \hat{V}_d^{C_a}(s)}}{VOO Regret bound} &\leq \mathcal{R}_{C_a}.
\end{align}
Thus, with our choice of $C_a$ that is designed to satisfy the recurrence relation, we obtain the bound:
\begin{align}
  \lra{V_d^\star(s) - \hat{V}_d^{C_a}(s)}&\leq \undb{\lra{V_d^\star(s) - \hat{V}_d(s)}}{VOO Recursive bound} + \undb{\lra{\hat{V}_d(s) - \hat{V}_d^{C_a}(s)}}{VOO Regret bound}\\
  &\leq \gamma \cdot \eta_{C_a}(d+1) + \mathcal{R}_{C_a} \leq \eta_{C_a}(d).
\end{align}

\paragraph{Step 2: Sparse Sampling Bound}
Similar to Lemma \ref{lemm:app-vowss}, the SS bound can also be further decomposed into the following terms:
\begin{align}
  \lra{\hat{V}_d^{C_a}(s) - \hat{V}_{\textsc{VOSS},d}(s)} &\leq \undb{\lra{\hat{V}_d^{C_a}(s) - \tilde{V}_{\textsc{VOSS},d}(s)}}{SS Concentration bound} + \undb{\lra{\tilde{V}_{\textsc{VOSS},d}(s) - \hat{V}_{\textsc{VOSS},d}(s)}}{SS Recursive bound}.
\end{align}
Here, the extra intermediate random variable and the VOSS estimator are defined in the following manner:
\begin{align}
  \tilde{V}_{\textsc{VOSS},d}(s) &\equiv \max_{a\in VOO(A,C_a)}\tilde{Q}_{\textsc{VOSS},d}(s,a)= \max_{a\in VOO(A,C_a)}\lrc{R(s,a) + \gamma \frac{1}{C_s}\sum_{i=1}^{C_{s}}\hat{V}_{d+1}^{C_a}(s_i')},\\
  \hat{V}_{\textsc{VOSS},d}(s) &\equiv \max_{a\in VOO(A,C_a)}\hat{Q}_{\textsc{VOSS},d}(s,a)= \max_{a\in VOO(A,C_a)}\lrc{R(s,a) + \gamma \frac{1}{C_s}\sum_{i=1}^{C_{s}}\hat{V}_{\textsc{VOSS},d+1}(s_i')}.
\end{align}
We now apply the sparse sampling bound as well as the recursive bound in order to bound the SS bound components. 
In our case, since our intermediate sparse sampling term $\tilde{V}_{\textsc{VOSS},d}(s)$ is merely swapping out the expectation of $\hat{V}_{d+1}^{C_a}$ with a sample average under the appropriate sampling density, this turns out to be simply the Chernoff bound.
We transform the sparse sampling concentration bound to an expected value bound by using the fact that the difference of $V$ functions/estimators is bounded above by $2V_{\max}$, setting the concentration bound to be $\epsilon$ as per the Lemma statement and assigning the worst case result with probability $p$. 
Once again, we compare the two quantities by picking a reference action to directly compare the quantities inside the max operations:
\begin{align}
  \undb{\lra{\hat{V}_d^{C_a}(s) - \tilde{V}_{\textsc{VOSS},d}(s)}}{SS Concentration bound} &\leq \lra{\max_{a\in VOO(A,C_a)}\{R(s,a) + \gamma \E_{s'\sim T(s,a)}[\hat{V}_{d+1}^{C_a}(s')]\} -\max_{a\in VOO(A,C_a)}\{R(s,a) + \gamma \frac{1}{C_s}\sum_{i=1}^{C_{s}}\hat{V}_{d+1}^{C_a}(s_i')\}}\\
  &\leq \gamma\lra{\E_{s'\sim T(s,a^*)}[\hat{V}_{d+1}^{C_a}(s')] - \frac{1}{C_s}\sum_{i=1}^{C_{s}}\hat{V}_{d+1}^{C_a}(s_i')} \tag{$a^* = \argmax_{VOO} \hat{Q}_d(s, a)$}\\
  &\leq \gamma \cdot ((1-p)\epsilon + 2p\cdot V_{\max}) \leq \gamma \cdot (\epsilon + 2p\cdot V_{\max}).
\end{align}

For the SS Recursive bound, we proceed with the similar recursive calculation as SS Recursive bound term done in Lemma \ref{lemm:app-vowss} by using the inductive hypothesis for step $d+1$: 
\begin{align}
    \undb{\lra{\tilde{V}_{\textsc{VOSS},d}(s) - \hat{V}_{\textsc{VOSS},d}(s)}}{SS Recursive bound} &\leq \ilim{}{}\max_{a\in VOO(A,C_a)}\lrc{R(s,a) + \gamma \frac{1}{C_s}\sum_{i=1}^{C_{s}}\hat{V}_{d+1}^{C_a}(s_i')}\\ 
    &\;\;\;\;\;\;\;\; -\max_{a\in VOO(A,C_a)}\lrc{R(s,a) + \gamma \frac{1}{C_s}\sum_{i=1}^{C_{s}}\hat{V}_{\textsc{VOSS},d+1}(s_i')}\ilim{}{}\\
    &\leq \gamma \frac{1}{C_s}\sum_{i=1}^{C_{s}}\lra{\hat{V}_{d+1}^{C_a}(s_i') - \hat{V}_{\textsc{VOSS},d+1}(s_i')} \tag{$\tilde{a} = \argmax_{VOO} \tilde{Q}_{\textsc{VOSS},d}(s, a)$}\\
  &\leq \gamma \cdot \alpha_{C_s}(d+1).
\end{align}
Putting the sparse sampling components together by applying the recursive definition of $\alpha_{C_s}(d)$, we obtain the recurring concentration inequality by induction:
\begin{align}
  \lra{\hat{V}_d^{C_a}(s) - \hat{V}_{\textsc{VOSS},d}(s)} &\leq \undb{\lra{\hat{V}_d^{C_a}(s) - \tilde{V}_{\textsc{VOSS},d}(s)}}{SS Concentration bound} + \undb{\lra{\tilde{V}_{\textsc{VOSS},d}(s) - \hat{V}_{\textsc{VOSS},d}(s)}}{SS Recursive bound}\\
  &\leq \gamma \cdot (\epsilon + 2p\cdot V_{\max}) + \gamma \cdot \alpha_{C_s}(d+1) = \alpha_{C_s}(d).\done
\end{align}

\textit{Proof for Theorem \ref{thm:app-voss}}. Finally, we prove the main theorem by combining the two terms:
\begin{align}
  \lra{V_d^\star(s) - \hat{V}_{\textsc{VOSS},d}(s)} &\leq \undb{\lra{V_d^\star(s) - \hat{V}_d^{C_a}(s)}}{VOO bound} + \undb{\lra{\hat{V}_d^{C_a}(s) - \hat{V}_{\textsc{VOSS},d}(s)}}{SS bound} \leq \eta_{C_a}(d) + \alpha_{C_s}(d) \leq \eta_{C_a} + \alpha_{C_s}. \done
\end{align}

Here, the $\eta_{C_a}(d)$ corresponds to the VOO bound in \cite{kim2020}.
Thus, we can always find a corresponding $C_a$ for an arbitrary decreasing sequence of $\eta_{C_a}(d)$ which satisfies the properties in our lemma, and this sequence can be made closer to 0 for each depth by choosing bigger $C_a$ values.

On the other hand, $\alpha_{C_s}(d)$ corresponds to the sparse sampling bound/Chernoff bound which decreases in both $p$ and $\epsilon$ with more samples $C_s$. Thus, it can also be made to decrease to 0 as we increase $C_s$.

\end{document}